%% file: acl_latex.tex
\pdfoutput=1

\documentclass[11pt]{article}

\usepackage[]{acl}

\usepackage{times}
\usepackage{latexsym}

\usepackage[T1]{fontenc}

\usepackage[utf8]{inputenc}

\usepackage{amssymb}
\usepackage{bbm}
\usepackage{bm}
\usepackage{titlesec}
\usepackage{multirow}
\usepackage{mathtools}
\usepackage{colortbl}
\usepackage{float}
\usepackage{xcolor}
\usepackage{adjustbox}
\usepackage{enumitem}
\usepackage{arydshln}
\usepackage[capitalize]{cleveref}

\usepackage{booktabs}
\usepackage[font=small,labelfont=bf,tableposition=top]{caption}
\DeclareCaptionLabelFormat{andtable}{#1~#2  \&  \tablename~\thetable}

\usepackage{pict2e}

\newcommand{\circled}[1]{%
  \setlength{\unitlength}{1em}%
  \begin{picture}(1,1)%
    \put(0.5,0.5){\circle{1}}%
    \put(0.5,0.5){\makebox(0,0){#1}}%
  \end{picture}%
}

\usepackage{microtype}

\usepackage{inconsolata}

\title{AMERICANO: Argument Generation with Discourse-driven Decomposition and Agent Interaction}

\author{Zhe Hu$^{1,2}$ \,  Hou Pong Chan$^{3}\thanks{~~Work was done while Hou Pong was at the University of Macau.}$   \, Yu Yin$^{4}$
\\
$^{1}$The Hong Kong Polytechnic University
\quad$^{2}$InspireOmni AI \\
  $^{3}$DAMO Academy, Alibaba Group
 \quad $^{4}$Case Western Reserve University
 \\
  $^{1}${\tt zhe-derek.hu@connect.polyu.hk},
  $^{3}${\tt houpong.chan@alibaba-inc.com}, 
  $^{4}${\tt yu.yin@case.edu}
  }

\begin{document}
\maketitle

\begin{abstract}
Argument generation is a challenging task in natural language processing, which requires rigorous reasoning and proper content organization. 
Inspired by recent chain-of-thought prompting that breaks down a complex task into intermediate steps, we propose \textsc{Americano}, a novel framework with agent interaction for argument generation.
Our  approach decomposes the generation process into sequential actions grounded on argumentation theory, which first executes actions sequentially to generate argumentative discourse components, and then produces a final argument conditioned on the components.
To further mimic the human writing process and improve the left-to-right generation paradigm of current autoregressive language models, we introduce an argument refinement module that automatically evaluates and refines argument drafts based on feedback received. We evaluate our framework on the task of counterargument generation using a subset of Reddit/CMV dataset. The results show that our method outperforms both end-to-end and chain-of-thought prompting methods and can generate more coherent and persuasive arguments with diverse and rich contents.
\end{abstract}

\section{Introduction}

\input{intro}

\section{Argument Generation with Discourse-driven Sequential Actions}
\input{method}

\section{Experiment Setup and Evaluation}
\input{experiment_setup}

\section{Results and Analysis}

\input{result_analysis}

\section{Related Work}
\input{related}

\section{Conclusion}
\input{conclusion}

\section*{Limitations}
Argument generation is a challenging task in natural language processing. In this work we propose a multi-agent based framework utilizing LLMs for counterargument generation. However, there are several limitations of our work. First, in our system, the refinement module only revises the argument draft without directly modifying the actions in the generation agent (i.e., claim, reasoning, concession). The feedback can be incorporated to further improve actions for initial argument draft generation. 
Second, debating is an interactive process where two sides can interactively debate with each other in a conversational way. Future work might study interactive argumentation with multiple debating agents. 
Third, our in-depth analysis shows that the system occasionally generates arguments with unverified or speculative evidence. Such instances highlight a critical area for future improvement, specifically the integration of fact-checking methods to enhance the reliability of the generated arguments.
Finally, in our experiments, GPT-3.5 is used as the base model. However, other LLMs, particularly smaller models (e.g., 7B and 13B models), can also be incorporated to further showcase the effectiveness of our framework.

\section*{Ethics Statement}
We recognize that our framework may generate fabricated and potentially harmful content due to the systematic biases of pre-training using heterogeneous web corpora and the open-ended generation characteristics of the argumentative text generation tasks.  
As our method utilizes large language models and does not require model training, the generated outputs may contain harmful and biased contents as the generation of language models can not be fully controlled. 
Argument generation is an open-ended generation task with objective opinions. Therefore, we urge the users to carefully examine the ethical influence of the generated outputs and cautiously apply the system in real-world applications. 

\bibliography{anthology,custom}

\clearpage
\newpage
\appendix
\input{appendix}




\end{document}

%% file: intro.tex
Argument generation is an essential task in natural language processing with wide applications, such as debates and essay writing~\cite{toulmin2003uses}. In this work, we study \textit{counterargument generation} which aims to generate persuasive arguments to refute a given proposition on a controversial topic~\cite{hua-wang-2018-neural,alshomary-wachsmuth-2023-conclusion}. However, generating counterarguments poses significant challenges for both humans and machines
as it requires a profound comprehension of the original proposition, the ability to present a valid standpoint from an opposing perspective, and the competence to provide rigorous reasoning to justify the claim~\cite{antaki1999show,grote1997ma,walton2008argumentation,wang-etal-2017-winning}. 


\begin{figure}[t]
\centering\includegraphics[width=\columnwidth,trim=0 0 0cm 0, clip]{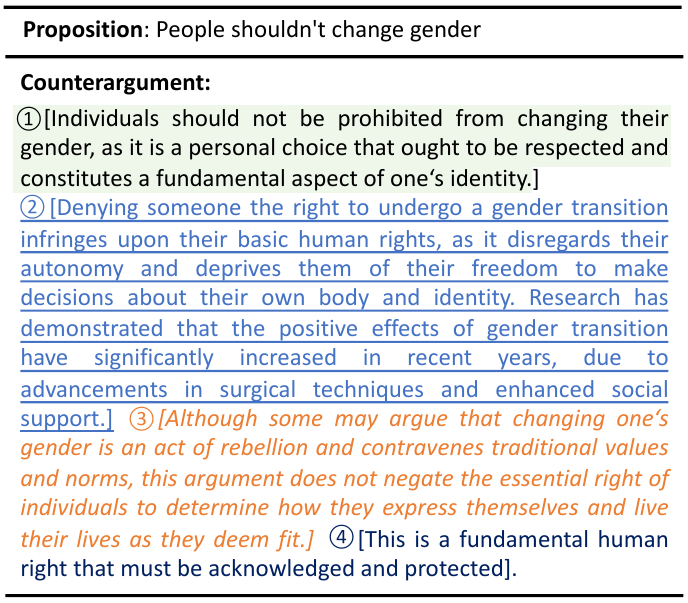}
    \captionof{figure}{ 
    Sample counterargument that refutes the proposition. The argument structure consists of components including \circled{1} a \textit{claim} serving as the main statement to attack the proposition, \circled{2} a \textit{reasoning} that supports the claim, \circled{3} a \textit{concession}  responding with potential rebuttals and \circled{4} a \textit{conclusion}. 
    }
    \vspace{-6mm}
    \label{fig:sample_intro}
\end{figure}

Recent large language models (LLMs) have exhibited remarkable capabilities in addressing various tasks with human-alike result~\cite{NEURIPS2020_1457c0d6,ouyang2022training,openai2023gpt4,chowdhery2022palm}. However, the token-level autoregressive generation paradigm makes LLMs fall short of dealing with complicated tasks involving multiple actions due to the lack of \textit{high-level planning} ability~\cite{bubeck2023sparks}. Prior work shows that chain-of-thought (CoT) prompting can significantly boost the LLMs' ability on complex reasoning tasks by encouraging the model to decompose the task into a sequence of intermediate results~\cite{wei2022chain}. Later work further imposes automatic decision-making and action-executing to break down complex tasks leveraging LLMs~\cite{shinn2023reflexion,yao2022react,sun2023adaplanner}. 

Although the above methods achieve good performance in solving reasoning tasks,  they still face challenges when applied to argument generation. Generating arguments \textit{not only requires rigorous reasoning but also demands deliberate discourse structures to enhance overall coherence and persuasion}~\cite{musi2018changemyview,hua-wang-2020-pair}. As shown in Figure~\ref{fig:sample_intro}, a counterargument comprises several discourse components, and
generating a strong argument needs both to produce high-quality components and to properly organize the components to ensure overall quality. Nevertheless, decomposing the goal of argument generation into intermediate actions remains a non-trivial task. Moreover, \textit{the left-to-right single-pass generation paradigm of current LLMs hinders them from tracking back and revising in previously generated text}. This limitation potentially depletes the soundness and coherence of the generated argument~\cite{wang-etal-2018-paper,madaan2023self,hu-etal-2022-mocha}. 


In this work, we propose \textbf{\textsc{Americano}}, a novel framework for \underline{a}rgu\underline{m}ent gen\underline{er}ation with d\underline{i}scourse-driven de\underline{c}omposition and \underline{a}gent i\underline{n}teracti\underline{o}n,
where a \textbf{generation agent} first produces an argument draft, and then an \textbf{evaluation agent} and \textbf{refinement agent} iteratively produce feedback and revise the draft. 
Inspired by argumentation theory and argumentative discourse structure~\cite{van2004systematic,green2010representation,palau2009argumentation}, our argument generation agent decomposes the goal into predefined actions and sequentially generates each argumentative discourse component. 
Specifically, given a proposition and the goal of generating a counterargument, the sequential actions aiming to create high-quality discourse components include:
(1) 
a \textit{claim action} that produces a strong claim to refute the proposition; (2) a \textit{reasoning action} that generates and revises a detailed logical reasoning to support the claim; (3) a \textit{concession action} that creates potential acknowledgements of the original proposition. Following the generation of these intermediate discourse components, an \textit{argument generation action} is executed to organize the intermediate contents and generate a final counterargument.


To further mitigate the drawback of left-to-right generation and incorporate feedback,
we propose an argument refinement module with two agents - an evaluation agent and a refinement agent.
Specifically, the argument draft is first evaluated by the evaluate agent to provide verbal feedback signals, and then the feedback is passed to the refinement agent to revise the draft. This process can be conducted iteratively until the evaluator is satisfied with the result.
Both agents are operated by prompting LLMs without any model training.
This is also akin to the human writing process of first composing a draft and then revising the draft~\cite{flower1981cognitive} to improve the quality.

We evaluate our framework on the task of zero-shot counterargument generation, with a subset of propositions collected from Reddit/CMV dataset. We leverage both LLM-based automatic evaluation and human evaluation to validate the model outputs. The results show that our method is able to produce high-quality counterarguments with better coherence and persuasiveness compared with end-to-end prompting and CoT prompting. Moreover, our system can generate more diverse results than baseline methods. Data and Code are available at: \url{https://github.com/Derekkk/LLM4ArgGen}.

%% file: method.tex
\begin{figure*}[t]
    \centering
    \includegraphics[scale=0.7]{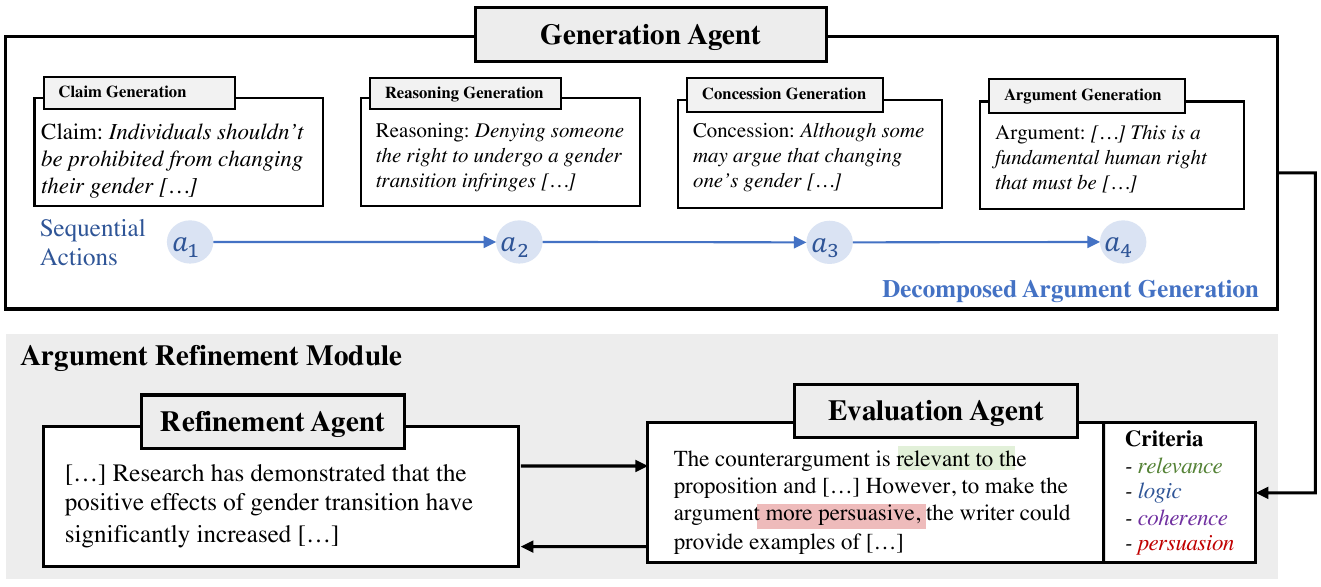}
    \vspace{-2mm}
    \captionof{figure}{Overview of our framework. The generator first decomposes the task into a sequence of actions and produces an initial result. Then, a refinement module with two agents iteratively provides feedback and revises the result.
    }
    \vspace{-4mm}
    \label{fig:overall}
\end{figure*}

The overall framework is shown in Figure~\ref{fig:overall}, which consists of three agents that collaboratively perform task decomposition and refinement for argument generation. We first introduce the generation agent. 

Argument generation can be modeled as $p(y|x)$, where $x$ is an input proposition and $y$ is an output counterargument. However, directly modeling this probability presents significant challenges, as generating arguments necessitates appropriate high-level planning, rigorous reasoning, and proper content organization.
Instead of directly prompting LLMs for argument generation, we decompose the goal into a sequence of actions based on argumentative discourse structure~\cite{stab-gurevych-2014-identifying,madnani-etal-2012-identifying,wambsganss-niklaus-2022-modeling}.
Each action tackles a subproblem based on the internal structure of an argument, which typically includes: a \textbf{claim} as the central statement the writer is trying to argue, a \textbf{reasoning} to support the claim, and an optional \textbf{concession/acknowledgement} to address potential dissenters and improve persuasion.~\footnote{We do not explicitly include a conclusion as the main claim can often be restated as the conclusion.} 

Driven by this, we break down the generation into sequential actions that first generate the components and then produce a final argument: $p(y|x)=p(y|a,r,c,x)p(a|r,c,x)p(r|c,x)p(c|x)$, where $c$, $r$, $a$ denotes claim, reasoning and acknowledge/concession respectively. Such modeling  reduces the complexity of $p(y|x)$. 
All the actions are conducted by prompting the same LLM ($\mathcal M$), eliminating the costly model training.



\begin{figure}[t]
    \centering
    \includegraphics[scale=0.52]{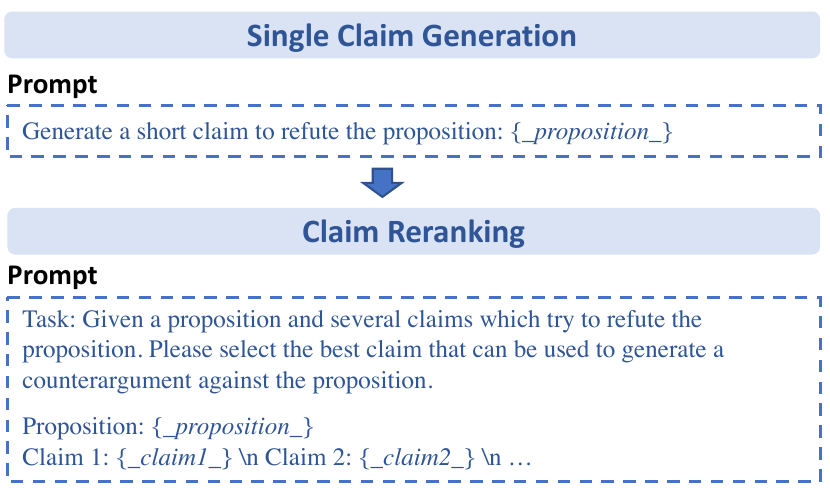}

    \captionof{figure}{ 
    Prompts for claim generation.
    }
    \vspace{-5mm}
    \label{fig:claim_gen}
\end{figure}

\subsection{Claim Generation Action}
The claim is the central component of an argument. For counterarguments, it should express a different stance regarding the proposition. As shown in Figure~\ref{fig:claim_gen}, we prompt $\mathcal{M}$ to generate a potential claim. However, multiple valid claims may exist given an input proposition.
Therefore, instead of executing the action only once, we prompt $\mathcal{M}$ multiple times to produce a set of claims and then introduce a claim reranking step to select the best one.

For claim reranking, we again utilize $\mathcal{M}$ to rank the claims based on the potential to generate a persuasive argument. To reduce variance and improve the self-consistency of the ranking, we further introduce a majority voting strategy by prompting $\mathcal{M}$ multiple times and selecting the claim that is ranked as topmost with the highest frequency. This simple strategy has proven effective in other tasks such as CoT prompting~\cite{wang2022self}.


\begin{figure}[t]
    \centering
    \includegraphics[scale=0.52]{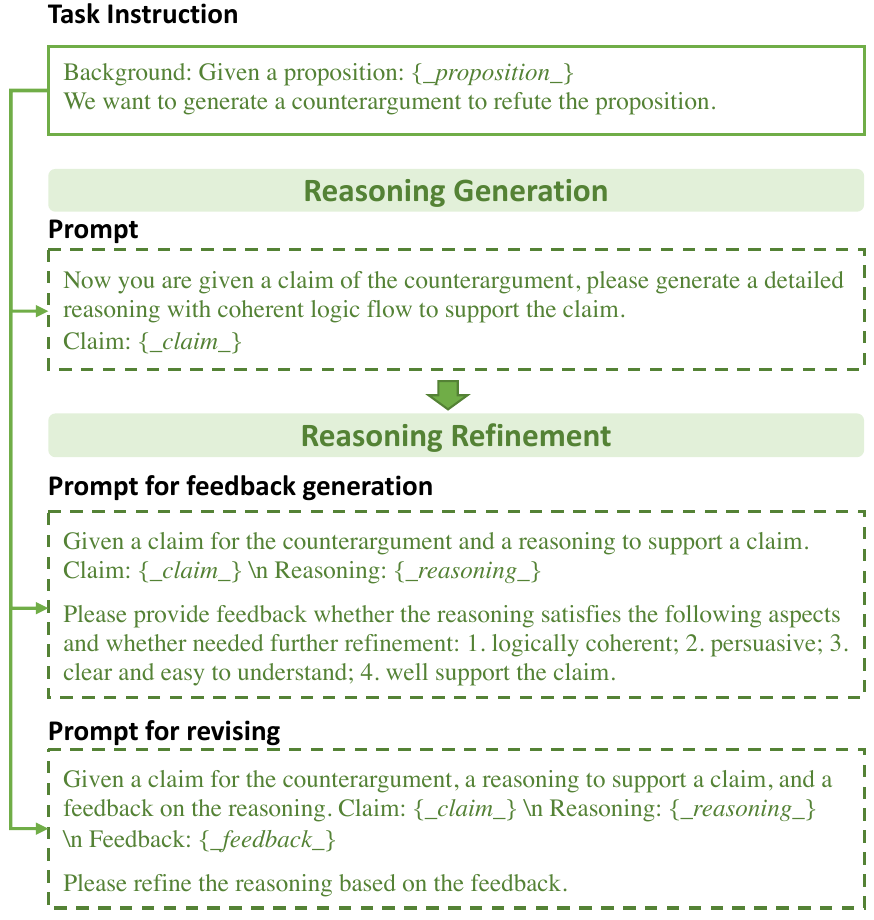}
    \vspace{-3mm}
    \captionof{figure}{ 
    Prompts for reasoning generation.
    }
    \vspace{-3mm}
    \label{fig:reason_gen}
\end{figure}

\subsection{Reasoning Generation Action}

Reasoning generation action aims to produce a comprehensive reasoning conditioned on both proposition ($x$) and the previously generated claim ($c$). As illustrated in  Figure~\ref{fig:reason_gen}, we first employ $\mathcal{M}$ to create an initial reasoning using the concatenation of the task instruction and prompt. Additionally, we leverage an off-the-shelf NLI model\footnote{\url{https://huggingface.co/MoritzLaurer/DeBERTa-v3-large-mnli-fever-anli-ling-wanli}} to verify that  generated reasoning entails the claim.

However, generating high-quality reasoning requires strict logical inference and internal consistency, which is difficult to achieve by only prompting LLMs once. Therefore, we leverage $\mathcal{M}$ as a critic to provide feedback and reinforce the generator to progressively revise the reasoning. 
We employ pre-defined criteria as verbal prompts, addressing aspects including logical coherence, persuasiveness, and whether the reasoning makes sense and well supports the claim. The generator then modifies the reasoning by additionally consuming the feedback. 
This process is conducted iteratively until no feedback is required or the maximum number of iterations is reached. 
This ensures  a strong reasoning is generated, which can be utilized to enhance the subsequent counterargument generation. 

\subsection{Concession Generation Action}

Concessions are considered as an argumentative strategy that enhances persuasion in discourse studies~\cite{mann1988rhetorical,musi2018changemyview,antaki1999show,wolfe2009argumentation}.
A concession, or acknowledgement, is typically employed to produce trust and fortify
one's position by addressing potential dissenters in an argument. 

This action aims to generate a concession based on the proposition, the previously generated claim, and reasoning. Similarly, we utilize $\mathcal{M}$ for concession generation.
As the concession should not weaken the original counterargument, we include the following instruction in the prompt:

\textit{{"Note that the goal of the concession is not to weaken the claim and reasoning, but to produce trust and make the counterargument more convincing and persuasive to the audience."}}

This instruction has proven effective in our initial experiments. The full prompt is in the Appendix~\ref{sec:detail_prompts}.

\begin{figure}[t]
    \centering
    \includegraphics[scale=0.52]{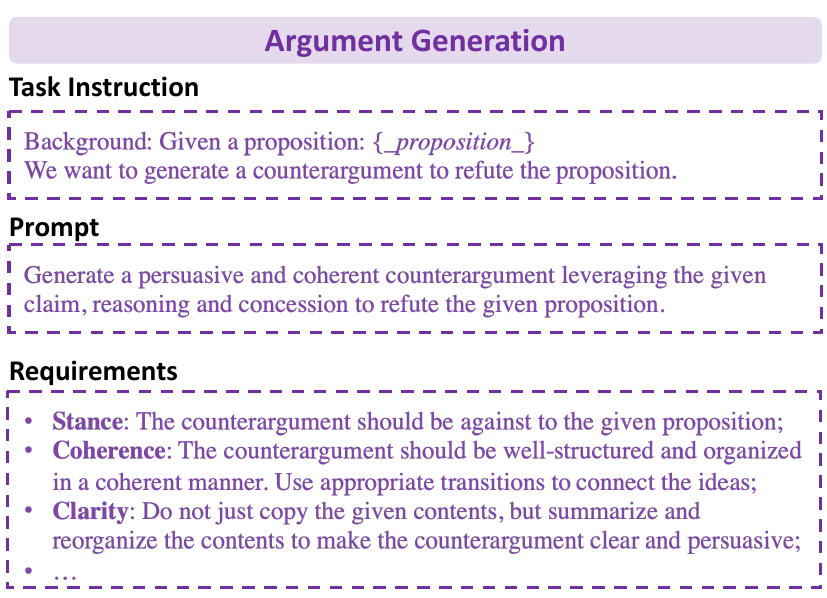}
    \vspace{-3mm}
    \captionof{figure}{ 
    Prompts for argument generation.
    }
    \vspace{-6mm}
    \label{fig:arg_gen}
\end{figure}

\subsection{Counterargument Generation Action}
Thus far, we have generated all the essential components of an argument, including a claim, reasoning and concession. 
Next, we generate the final counterargument based on these components. This step requires properly understanding the components and effectively organizing the content to produce a coherent outcome. We again rely on $\mathcal{M}$  to execute the action. As shown in Figure~\ref{fig:arg_gen}, besides the task instruction and prompt, we further include pre-defined requirements on aspects including stance, coherence, and clarity, to enhance overall performance and effectiveness.

\section{Argument Refinement Module}
Previous work has shown that producing an output 
on a single attempt is challenging for both machines and human beings
~\cite{hua-wang-2020-pair,hu-etal-2022-mocha,wang-etal-2018-paper}. Conventional autoregressive language models produce outputs from left to right at the token level, lacking the capacity to edit and revise previously generated content. Drawing inspiration from the human writing process that involves first creating an initial draft and subsequently refining it, we propose an argument refinement module to mimic this process. As illustrated in Figure~\ref{fig:overall}, this module comprises an evaluation agent and a refinement agent. The evaluator first provides feedback on the current draft, and then the refinement agent takes the feedback and revises the draft. Two agents interact with each other interatively to formalize an optimization process for generation.

\subsection{Evaluation Agent for Feedback Generation}
Given a proposition and an initial counterargument draft, the evaluation agent first provides feedback on improving the counterargument. First, a valid counterargument should possess an opposing stance compared with the original proposition, and hence we leverage the same NLI model as in the reasoning generation to compute the relationship class $s_{arg}$ between the proposition and the counterargument. This result will be used in later steps if the predicted label does not correspond to \textsc{"contradiction"}. 

Furthermore, we leverage $\mathcal{M}$ to assess the counterargument draft and generate feedback. 
The evaluation criteria for counterargument include aspects of \textbf{relevance}, \textbf{logical consistency}, \textbf{coherence of structure}, and \textbf{persuasion}. These elements are fundamental aspects for constructing a solid argument. In future work, we plan to explore the integration of additional aspects into the refinement module. The detailed prompts can be found in the Appendix~\ref{sec:detail_prompts}.


\subsection{Refinement Agent}
The refinement agent takes as input the feedback and generates a revised version of the counterargument in each iteration. Concretely, it first verifies the stance based on the prediction of the NLI model: if the NLI label is not \textsc{"contradiction"}, it first utilizes $\mathcal{M}$ to adjust the draft so that its stance aligns with a valid counterargument that attempts to refute the original proposition, with the prompt: 
\textit{"The stance is wrong. The counterargument should be against the statement."}. 
Subsequently, it refines the counterargument by addressing the feedback from the evaluator to enhance the overall quality. 
The two agents work together in a loop until the evaluator is satisfied with the result. In practice, we bound the process by a maximum number of iterations.

Our refinement module  distinguishes itself from Self-refine~\cite{madaan2023self} in the way that they leverage the same LLM instance to serve as the generator, evaluator, and revisor, without any task decomposition. In contrast, our generation agent features a sequence of actions designed to produce high-quality initial results, offering a superior starting point for the refinement process, ultimately resulting in enhanced efficiency and effectiveness.

%% file: experiment_setup.tex
\subsection{Task Setup}
We evaluate our framework on the task of counterargument under a zero-shot setting, where the model is asked to generate a counterargument to refute a given proposition on a controversial topic. We randomly sample 50 propositions from Reddit/CMV dataset~\cite{hua-etal-2021-dyploc,hu-etal-2022-planet}, which is a counterargument generation dataset with samples collected from Reddit/ChangeMyView. All propositions are in the politics and policy domains. The full list of input propositions are in Table~\ref{fig:input_query}.

\subsection{Model Implementations and Baselines}
As we study \textit{zero-shot} argument generation, we compare our model with recent instructional LLMs.
We use GPT-3.5 (text-davinci-003) as the base LLM. 
We consider the baselines: (1) End-to-end generation (E2E) which directly prompts the LLM to generate a counterargument without any intermediate steps; (2) Plan-based CoT generation (PlanCoT) that first generates a chain of plans as intermediate content planning, and then produces the counterargument based on the plan;
(3) Our model variant without refinement module. All the baseline models use the same GPT-3.5 version as our framework. More details are in Appendix~\ref{sec:exp_details}.

\subsection{Evaluation Metrics}
We employ both automatic and human evaluations in our experiments. 
Automatically evaluating open-ended text generation tasks is a challenging task~\cite{celikyilmaz2020evaluation}.
Recent work has shown that leveraging LLMs to conduct reference-free text generation evaluation 
aligns well with human preference~\cite{liu2023gpteval,fu2023gptscore}. Therefore, we propose a LLM-based counterargument evaluation method leveraging GPT-4~\cite{openai2023gpt4} to judge the ouptuts.~\footnote{We do not include reference-based metrics due to the open-ended nature of argument generation, where multiple valid arguments may exist for the same input.}

\begin{figure}[t]
    \centering
    \includegraphics[scale=0.53]{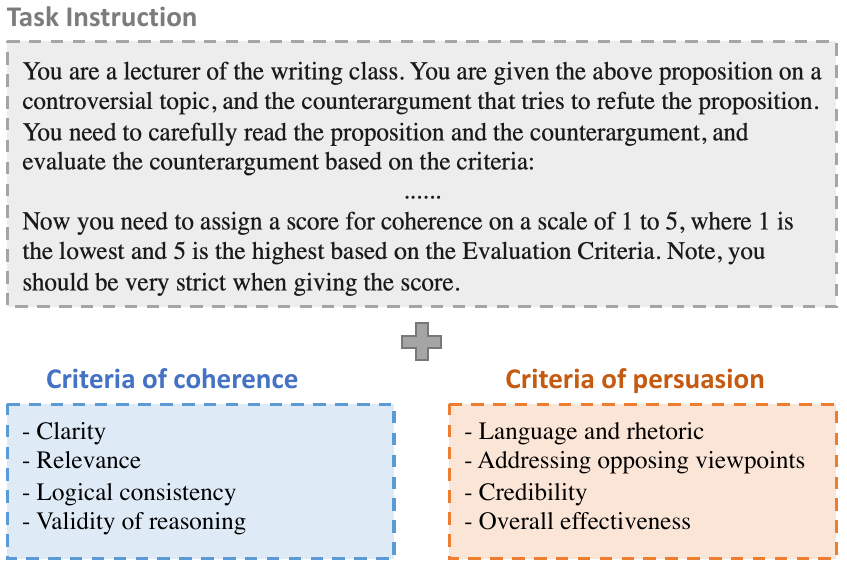}
    \captionof{figure}{ 
    LLM-based automatic evaluation.
    }
    \vspace{-5mm}
    \label{fig:gpt4_eval}
\end{figure}

\subsubsection{LLM-based Automatic Evaluation}
\label{sec:gpt4_auto_eval}
In our LLM-based evaluation, we focus primarily on two aspects: \textbf{coherence} and \textbf{persuasion}. These two aspects are essential elements of a good argument with clear definitions and criteria, making them well-suited for assessments based on LLMs. Concretely, we leverage GPT-4 to evaluate coherence and persuasion by scoring the outputs on a scale of 1 to 5, with the higher score signifying superior quality. To reduce randomness, we evaluate each sample 5 times and average the scores.

The prompts used for evaluation are designed with specific task instructions and a comprehensive list of detailed criteria, depicted in Figure~\ref{fig:gpt4_eval}. 
For \underline{coherence}, we concentrate on assessing both logical and discourse coherence, measuring the score jointly based on clarity, relevance to the proposition, logical consistency and soundness of reasoning. For \underline{persuasion}, we appraise the outputs according to language and rhetoric usage, the ability to address opposing viewpoints, credibility of evidence, and the overall effectiveness to persuade the audience. Each aspect comes with a detailed explanation. To improve stability, we prompt model to first generate a detailed rationale and then predict the score. More details are in Appendix~\ref{sec:gpt4_eval}.

\subsubsection{Human Evaluation}
For human evaluations, we hire three proficient English speakers as judges to evaluate output quality. Following prior research~\cite{hua-etal-2019-argument-generation,hua-etal-2021-dyploc}, we evaluate on the following aspects: \textbf{Appropriateness}-measures if an output is clear, readable and logical consistent;  \textbf{Content Richness}-represents the amount of informative talking points; and \textbf{Overall Quality}. Given an input proposition and several model outputs, the judges are asked to rank the outputs according to each aspect. In addition, we ask the judges to identify \textbf{Valid} counterarguments of high quality, focusing on the intrinsic merits of an output as a standalone, compelling argument, rather than its relative ranking against others. We select 30 random instances for evaluation. More details and the guidelines are in Appendix~\ref{sec:human_eval_details}.

%% file: result_analysis.tex
\begin{table*}[t]
\fontsize{10}{12}\selectfont
 \setlength{\tabcolsep}{1.5mm}
  \centering
    \begin{tabular}{lcc c c}
        \toprule 
        {\bf Method}  & {\bf Appropriateness (↑)} & {\bf Content Richness (↑)} & {\bf Overall Quality (↑)} & {\bf \% Validity (↑)} \\
        \midrule
        E2E  &  1.81 / 6.7\% & 1.54 / 0.0\% & 1.78  / 4.4\% & 45.00\%\\
        PlanCoT & 2.00 / 14.4\% & 1.74 / 6.7\%&  2.99 / 10.0\% & 53.33\% \\
        \hdashline
        $\text{Ours}_{\text{w/o Refine}}$ & 3.00 / 34.4\% & 3.16 / 32.2\% & 3.03 / \ 28.9\% &  \textbf{91.67\%} \\
        Ours & \textbf{3.12} / \textbf{44.4\%}  & \textbf{3.44}* / \textbf{61.1\%} &  \textbf{3.31}* / \textbf{56.7\%} &  {86.67\%} \\
        \bottomrule
    \end{tabular}
    \vspace{3mm}
    \caption{
    Human evaluation results. 
    For appropriateness, informativeness, and overall quality, the first score is computed based on the relative ranking position, and the second value represents the frequency of the output being ranked as the topmost. For validity, we present the percentage of results that are deemed to be generally strong arguments of high quality. (*: significantly better than all comparisons with p < 0.05, using Welch's t-test) 
  }
  \label{tab:human_result}
  \vspace{-4mm}
\end{table*}

\subsection{Automatic Results}
\begin{figure}[t]
    \centering
\centering\includegraphics[scale=0.34]{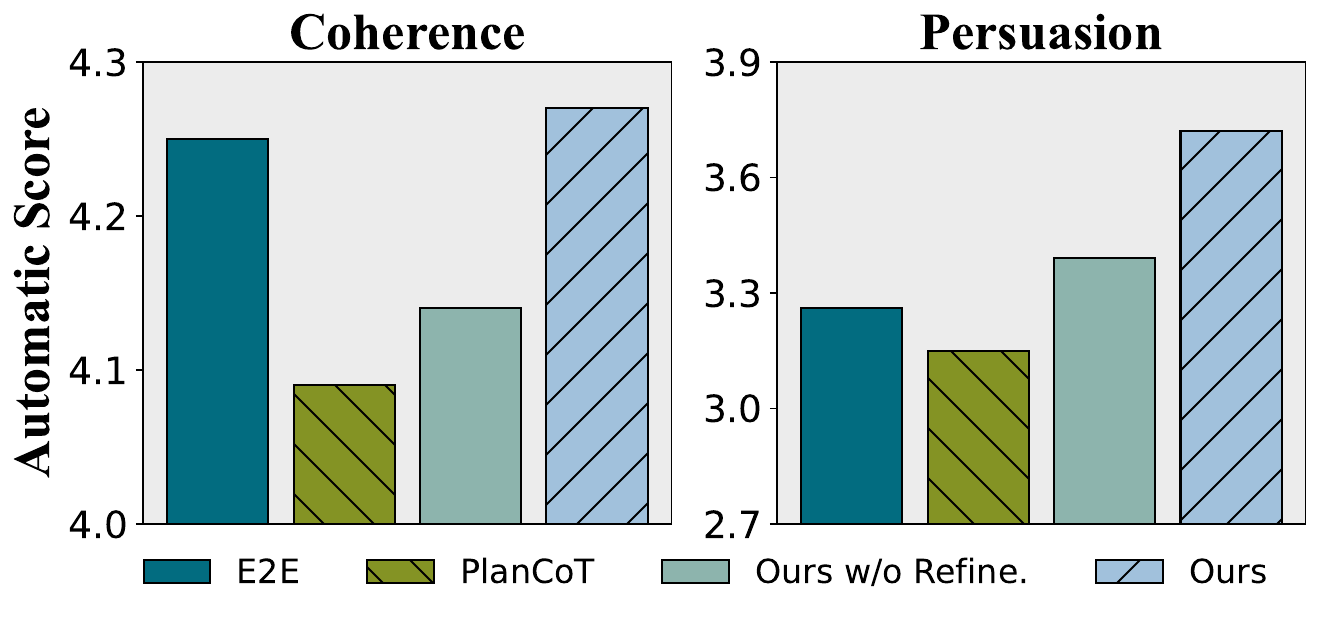}
    \vspace{-3mm}
    \captionof{figure}{ 
    Automatic results on coherence and persuasion by GPT4-based evaluation. A larger score means better quality. 
    }
    \vspace{-3mm}
    \label{fig:gpt4_pred_res}
\end{figure}

The LLM-based evaluation results on coherence and persuasion are displayed in Figure~\ref{fig:gpt4_pred_res}. As can be seen, our method outperforms all baselines in terms of persuasion and coherence, demonstrating the effectiveness of our framework in generating high-quality arguments. 

Specifically, for coherence, we observe that decomposing the generation (Ours w/o Refine) results in reduced coherence compared with E2E. One possible reason is that generating a final argument based on argumentative discourse components requires a deep understanding of each component and proper content organization, posing challenges when executing the argument generation action only once. 
Especially, our decomposed generation tends to produce longer outputs,~\footnote{We provide additional analysis on the impact of output length in Appendix~\ref{sec:length_analysis}.} further complicating the task of generating a coherent result in a single step.
However, incorporating the refinement module significantly boosts the coherence score, proving the importance of the refinement module in improving the overall coherence. 
For persuasion, both our model and the decomposed generation achieve higher scores compared to E2E and PlanCoT. The manual inspections show that our model outputs tend to include more talking points in the arguments, thus making the results more persuasive. This is further proved by the higher content richness scores of our model variants in human evaluations, as shown in Table~\ref{tab:human_result}.

\subsection{Human Evaluation Results}
The human evaluation results are shown in Table~\ref{tab:human_result}. As the evaluation of appropriateness, informativeness and overall quality are ranking-based, we convert the ranks into scores determined by subtracting its position in the ranking from the total number of candidates, with higher scores indicating better quality. Given that there are four models evaluated, the scores range from 1 (indicating the lowest rank) to 4 (indicating the highest rank). We also present the percentage of times the result is considered the top one for each aspect. 

First, $\text{Ours}_\text{w/o Refine}$'s results are ranked higher on all aspects compared with E2E and PlanCoT. This demonstrates that breaking down the E2E generation helps to maintain high-quality discourse components, ultimately leading to improved quality of final arguments. Second, the better content richness of our decomposed generation indicates that our model can produce outputs with more informative talking points to support the claim. This can be attributed to the reasoning generation action's ability to revise reasoning and make it stronger. Third, our discourse-driven sequential actions are more effective at improving the results compared with PlanCoT's content-based plans, making them better suited for argument generation.

By incorporating the refinement module to further reinforce the generation process, the results exhibit substantial improvements across all aspects except for validity. Our manual inspection by checking model outputs reveals that, on the one hand, the refinement module can reorganize argument content and reform the draft to achieve a better discourse structure, increasing the readability and coherence of the argument. On the other hand, during the refinement process, the model tends to add more detailed examples to support the claim, enhancing content richness and overall persuasion.

We also ask human judges to determine whether a generated result qualifies as a valid high-quality counterargument. While only 45\% of E2E generation results are considered valid, introducing CoT improves the outcomes, showing that decomposing complex goals is beneficial. $\text{Ours}_\text{w/o Refine}$ with discourse-driven actions achieves significantly better results, with almost 92\% of samples considered valid counterarguments, validating the effectiveness of incorporating discourse information into sequential actions. Interestingly, our full model with refinement scores approximately 87\% in validity, which is slightly lower than $\text{Ours}_\text{w/o Refine}$. The manual inspection reveals that, occasionally, the refinement process may introduce redundancies to the outputs and diminish overall readability, thus leading to a reduced validity. However, incorporating the refinement module is useful for ensuring that the generated arguments maintain the correct stance (i.e., refuting the input) and overall coherence, as direct generation may not always guarantee this. Further improvements to the refinement module are left for future work.

\subsection{Analysis on Output Diversity}
\label{subsec:diversity}

\begin{figure}[t]
    \centering
    \includegraphics[scale=0.3]{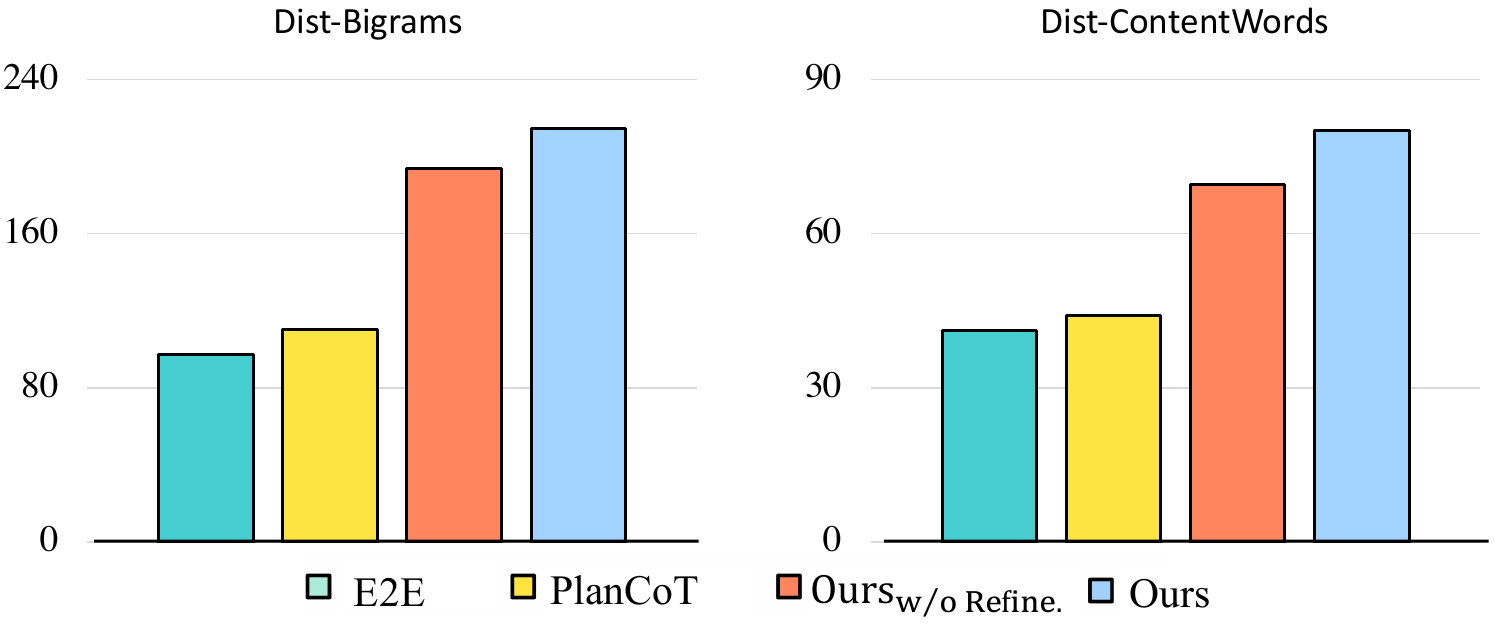}
    \vspace{-2mm}
    \captionof{figure}{ 
    Average number of distinct bigrams (Dist-Bigrams) and content words (Dist-ContentWords).
    }
    \vspace{-5mm}
    \label{fig:dist_res}
\end{figure}

We follow previous work and analyze output diversity by calculating the average number of distinct bigrams~\cite{li-etal-2016-diversity} and content words in each output. The results are in Figure~\ref{fig:dist_res}. Our method generates the most distinct bigrams and content words, demonstrating its ability to produce more diverse outputs. In contrast, E2E produces the least diverse outputs. We hypothesize that this may result from the fact that an LLM trained with RLHF tends to produce safer outputs without directly optimization for diversity.
After applying chain-of-thought, PlanCoT generates more diverse results. Leveraging our discourse-driven actions further improves scores, indicating that decomposing argument generation based on discourse components effectively enhances output diversity and content richness. We also present commonly generated verbs and nouns in Appendix~\ref{sec:word_cloud}.



\subsection{Further Analysis on Discourse}

\begin{figure}[t]
    \centering
    \includegraphics[scale=0.36]{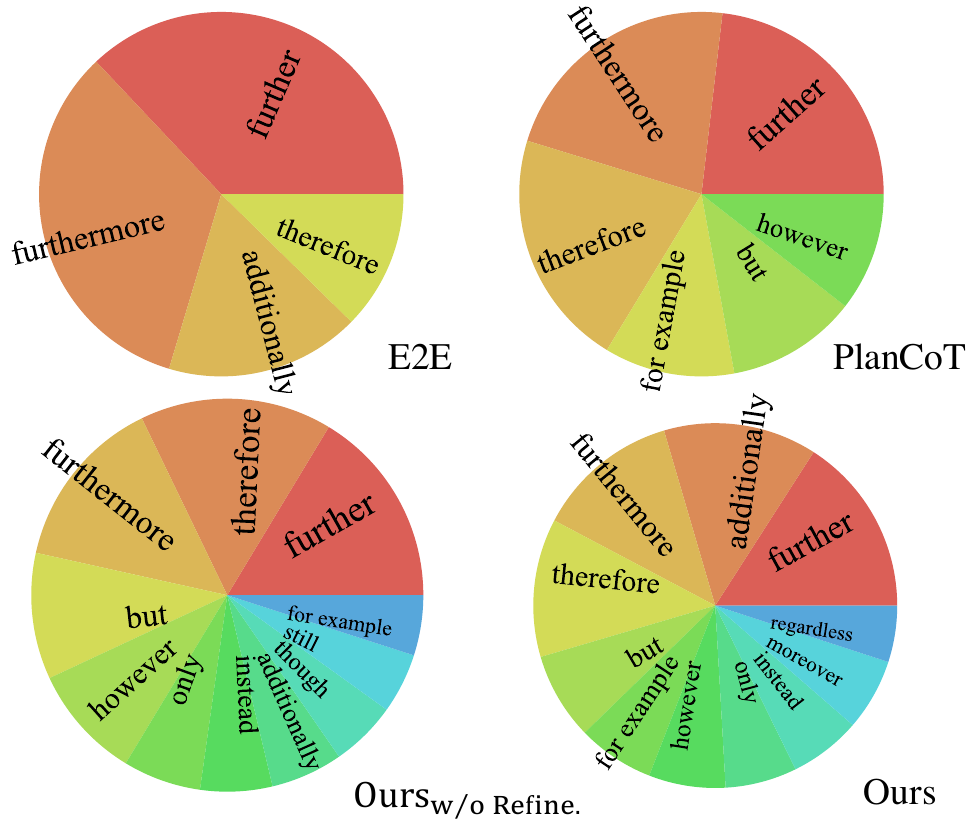}
    \vspace{-3mm}
    \captionof{figure}{ 
    Frequent discourse markers in outputs.
    }
    \vspace{-2mm}
    \label{fig:disco_markers}
\end{figure}



\input{sample_output_main}

\begin{figure}[t]
    \centering
    \includegraphics[scale=0.32]{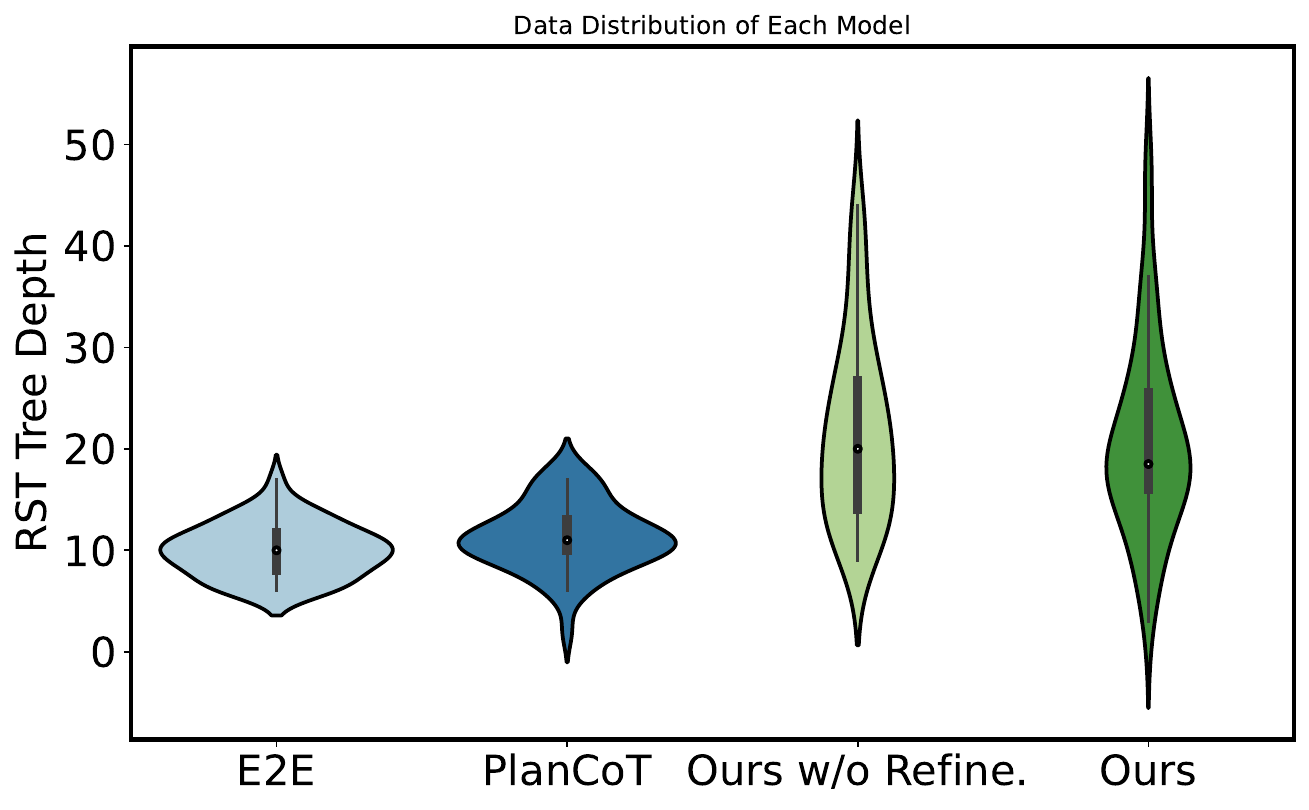}
    \vspace{-2mm}
    \captionof{figure}{ 
    Distribution of RST tree depth of  arguments.
    }
    \vspace{-3mm}
    \label{fig:rst_depth}
\end{figure}

\smallskip
\noindent\textbf{Usage of Discourse Markers.} 
Discourse markers are key features for modeling coherence~\cite{callaway-2003-integrating,grote-stede-1998-discourse} in various tasks~\cite{eckle-kohler-etal-2015-role,sharma-etal-2019-entity,samy-gonzalez-ledesma-2008-pragmatic}. We adopt the discourse markers from~\citet{sileo-etal-2019-mining} and extract frequent discourse markers that appear more than 10 times for each model, and present the results alongside their frequency in Figure~\ref{fig:disco_markers}. 

While only 4 common discourse markers are observed in the E2E results, applying chain-of-thought increases the usage of more markers. This is consistent with previous observations that chain-of-thought contributes to improved diversity. Moreover, both our methods, with or without refinement, leverage significantly more discourse markers. 
Compared with E2E and PlanCoT, our model variants employ discourse markers such as "though" and "regardless" because of the inclusion of concession components. Furthermore, the use of "for example" implies that our model variants learn to include more examples to support the claim, making the overall argument more persuasive.

\smallskip
\noindent\textbf{Analysis of RST Tree.} The discourse structure provides insight into the high-level organization of text. Following ~\citet{hua-wang-2020-pair}, we analyze the Rhetorical Structure Theory (RST) tree of the generated arguments. Concretely, we utilize an off-the-shelf discourse parsing tool~\cite{ji-eisenstein-2014-representation} to convert the arguments into RST trees and analyze the depth of the trees, illustrated in Figure~\ref{fig:rst_depth}. As can be seen, our model variants yield arguments with deeper structures. Moreover, a noticeable diversity in depth can be observed, spanning a broader spectrum compared to both E2E and PlanCoT models.
This further implies that our model can produce arguments with more diverse patterns and complicated structures.

\subsection{Sample Output Analysis}

We present a sample output with intermediate results in Figure~\ref{fig:samples1}. Given a proposition, our sequential actions first properly generate each discourse component and then organize them to form a coherent argument. Moreover, the final argument conforms to the discourse components and faithfully reflects each reasoning point, further indicating the strong controllability of the decomposed generation. It also implies our model's potential to be applied to \textit{interactive writing} wherein users could modify intermediate components and let the model organize the contents and generate final results.
We leave this to future work. Notably, during the refinement process, our model effectively adds \textit{examples in the concession part} to strengthen the whole argument. This proves that our refinement process can gradually improve the results. In addition, the underlined discourse markers in the final argument show that our model can properly utilize discourse markers and generate coherent outputs. More samples can be found in  Appendix~\ref{sec:addition_samples}.

%% file: sample_output_main.tex
\begin{figure*}[t]
    \bgroup
    \def\arraystretch{1.8}
	\fontsize{8}{8}\selectfont
     \hspace{-1mm}
	\setlength{\tabcolsep}{0.8mm}
	\centering
	\begin{tabular}{p{155mm}}
	\toprule
	\textbf{Proposition}: The fact that voting isn't mandatory is a good thing. \\
	\midrule	
        \vspace{-5mm}
        {\textbf{Claim}}: {\color{blue!75}Voting should be mandatory because it is a civic duty that ensures that the people's voices are heard and that their opinions are represented in government.}\\
        \vspace{-6mm}
        \textbf{Reasoning}: 
        [...]
        {\color{red!70}\textbf{{\scriptsize \circled{1}}}}Mandatory voting is a way of ensuring that all citizens are engaged in the electoral process and that their voices are taken into consideration when decisions are being made. {\color{green!70!blue}\textbf{{\scriptsize \circled{2}}}}It also helps to reduce the influence of special interests and ensures that the government truly reflects the will of the people. {\color{orange!65}\textbf{{\scriptsize \circled{3}}}}Furthermore, mandatory voting can help to reduce the influence of money in politics by ensuring that all citizens are represented and can participate in the democratic process. [...]
        \\
        \vspace{-6mm}
        \textbf{Concession}: {\color{teal}While it is true that mandatory voting can ensure that the voices of all citizens are heard, it is important to consider how to make the process accessible and fair for those who are unable to vote due to physical or financial limitations.}\\
        \hdashline
	{\textbf{Counterargument}}: The fact that voting isn't mandatory is not a good thing. {\color{red!70}\textbf{{\scriptsize \circled{1}}}}Mandatory voting should be implemented because it is a fundamental part of democracy and is an essential way to ensure that citizens are engaged in the electoral process and their voices are taken into consideration when decisions are made. {\color{green!70!blue}\textbf{{\scriptsize \circled{2}}}}\underline{By} requiring citizens to vote, it can help to reduce the influence of money in politics, as more people are participating in the democratic process and their opinions are being represented. {\color{orange!65}\textbf{{\scriptsize \circled{3}}}}\underline{Additionally}, mandatory voting can help to ensure that the government truly reflects the will of the people, as all citizens have an equal say in the political process. \\
    \vspace{-5mm}
    {\color{teal}\underline{However}, it is important to consider how to make the process accessible and fair for those who are unable to vote due to physical or financial limitations.} \textit{\underline{For example}, measures such as providing transportation to and from polling stations, offering absentee ballots, and introducing flexible voting hours can help to make the process easier and more accessible for those who have difficulty voting in person.} \underline{Additionally}, lawmakers should focus on creating laws that promote fairness and accessibility for all citizens, regardless of their socio-economic status. \\
    \vspace{-5mm}
    {\color{blue!75}\underline{Overall}, mandatory voting is a way to guarantee that all citizens have a voice in the political process and to ensure that the government is truly representative of the people's wishes}. \underline{By} implementing measures that make the voting process easier and more accessible for all citizens, mandatory voting can help ensure that everyone is given an equal chance to participate and have their voices heard\\
    \bottomrule
	\end{tabular}
    \vspace{-2mm}
\caption{Sample output of our method with the intermediate discourse components. 
} 
\label{fig:samples1}
\egroup
\vspace{-5mm}
\end{figure*}

%% file: related.tex
\textbf{Task Decomposition and Reasoning.} LLMs have achieved impressive results in solving various tasks with prompting~\cite{openai2023gpt4,brown2020language,anil2023palm,bubeck2023sparks}. However, the token-level left-to-right generation process limits the model's ability to tackle more complex tasks~\cite{bubeck2023sparks}.
To further improve model ability on complex reasoning tasks, recent work involves sampling intermediate reasoning steps~\cite{wei2022chain,nye2021show,wang2022self} or decomposing the complicated goal into actions~\cite{sun2023pearl,hao2023reasoning,zhou2022least,chen2023complex}. In this paper, we focus on the specific task of argument generation and decompose the goal into a sequence of predefined actions based on the argumentative theory to generate each discourse component.


\smallskip\noindent\textbf{Argument Generation.} Argument generation requires text planning, logical reasoning, and content organization~\cite{CARENINI2006925,hua-wang-2018-neural}. ~\citet{hua-etal-2019-argument-generation} propose a planning-based model with a retrieval module for counterargument generation. ~\citet{schiller-etal-2021-aspect} utilize keywords to control the content of arguments. ~\citet{bao-etal-2022-aeg} introduce a dual-decoder model to improve content planning. Different from previous work, we leverage LLMs and introduce a framework with multi-agents for counterargument generation. Our method effectively decomposes argument generation into subproblems and prompts LLMs for each action without model training.

\smallskip\noindent\textbf{Feedback and Refinement for Text Generation.} Previous work refines text generation by directly revising outputs without feedback~\cite{wang-etal-2018-paper,hu-etal-2022-mocha} or masking contents with low probability~\cite{hua-wang-2020-pair}. Recent work utilizes LLMs to provide feedback and reinforce language agents to improve model performance~\cite{shinn2023reflexion,sun2023adaplanner,madaan2023self,liang2023encouraging}. In this work, we introduce a refinement module with specifically designed criteria for argument refinement. Different from ~\citet{madaan2023self} which uses only one LLM instance for generation, evaluation, and refinement, our system consists of multiple agents that decompose generation with sequential actions, thus providing a better starting point for the refinement module and further encouraging divergent thinking of LLMs.

%% file: conclusion.tex
In this work, we present a novel framework for argument generation with agent interaction. Our framework consists of a generation agent that decomposes argument generation into a sequence of predefined actions driven by argumentative theory to produce a draft, and then a refinement module with an evaluator and a refinement agent to iteratively provide feedback and refine the draft. All parts are implemented leveraging LLMs with zero-shot prompting.
Both automatic and human evaluations show that our framework can generate more coherent and persuasive results with better diversity in counterargument generation.

%% file: appendix.tex
\newpage

\section{Experimental Details}
\label{sec:exp_details}
In our experiments, all modules of our methods and baselines are implemented by prompting GPT-3.5 (text-davinci-003)~\footnote{\url{https://platform.openai.com/docs/model-index-for-researchers}} as the based LLM. For hyper-parameters, we set temperature as 0.7 and top-p as 1, the maximum tokens are set as 2048.
For claim generation, we set the number of claims to be generated as 5. We set the maximum of iteration for refining reasoning and final argument as 3 and 1 respectively, considering the cost of API.
For automatic evaluation, we use GPT4 (gpt-4-0314)~\footnote{\url{https://platform.openai.com/docs/models/gpt-4}} as the base model.

\smallskip
\noindent\textbf{Discourse Markers.}
For the result analysis on discourse markers, we select markers from ~\citet{sileo-etal-2019-mining}. Some common markers such as "and", "or" are removed from the list. The complete list is presented in Figure~\ref{fig:disco_list}.

\begin{figure}[t]
    \def\arraystretch{1.5}
	\fontsize{8}{9}\selectfont
     \hspace{-2mm}
	\setlength{\tabcolsep}{0.8mm}
	\centering
	\begin{tabular}{|p{76mm}|}
    \hline
	absolutely, accordingly, actually, additionally, admittedly, afterward, again, already, alternately, alternatively, although, altogether, amazingly, anyway, apparently, arguably, as a result, basically, because of that, because of this, besides, but, by comparison, by contrast, by doing this, by then, certainly, clearly, coincidentally, collectively, consequently, conversely, curiously, currently, elsewhere, especially, essentially, eventually, evidently, finally, first, firstly, for example, for instance, fortunately, frankly, frequently, further, furthermore, generally, gradually, happily, hence, historically, honestly, hopefully, however, ideally, immediately, importantly, in contrast, in fact, in other words, in particular, in short, in sum, in the end, in the meantime, in turn, incidentally, increasingly, indeed, inevitably, initially, instead, interestingly, ironically, lastly, lately, later, likewise, locally, luckily, maybe, meaning, meantime, meanwhile, moreover, mostly, namely, nationally, naturally, nevertheless, next, nonetheless, normally, notably, obviously, occasionally, oddly, often, on the contrary, on the other hand, once, only, optionally, originally, otherwise, overall, particularly, perhaps, personally, plus, preferably, presently, presumably, previously, probably, rather, realistically, really, recently, regardless, remarkably, sadly, second, secondly, separately, seriously, significantly, similarly, simultaneously, slowly, sometimes, soon, specifically, still, strangely, subsequently, suddenly, supposedly, surely, surprisingly, technically, thankfully, then, theoretically, thereafter, thereby, therefore, third, thirdly, though, thus, together, traditionally, truly, truthfully, typically, ultimately, undoubtedly, unfortunately, unsurprisingly, usually, yet \\
	
    \hline
	\end{tabular}
    \vspace{-3mm}
\caption{List of discourse markers for result analysis. 
} 
\label{fig:disco_list}
\vspace{-2mm}
\end{figure}

\section{Detailed Prompts}
\label{sec:detail_prompts}
Here we provide detailed prompts for each module. Specifically, the prompt for concession generation is presented in Figure~\ref{fig:prompt_concession}. For the argument refinement module in our framework, the detailed prompt for the evaluation agent is presented in Figure~\ref{fig:prompt_refine_evaluator}, and the detailed prompt for the refinemnt agent in shown in Figure~\ref{fig:prompt_refine_revisor}.

The prompt for PlanCoT is presented in Figure~\ref{fig:prompt_cot}. For PlanCoT, we match the content of Counterargument as the final results, and do not use the plan in our experiments.

\begin{table}[t]
\fontsize{10}{12}\selectfont
 \setlength{\tabcolsep}{2.0mm}
  \centering
    \begin{tabular}{l c c}
        \toprule 
        {\bf Method}  & {\bf Coherence}  & {\bf Persuasion}\\
        \midrule
        E2E  &  3.84 & 3.64  \\
        PlanCoT & 3.83 & 3.58\\
        Ours & {\bf 3.87} & {\bf 3.78}
       \\
        \bottomrule
    \end{tabular}
    \vspace{2mm}
    \caption{
    GPT4-based automatic evaluations of generated arguments under length constraints.
  }
  \label{tab:automatic_results_300words}
  \vspace{-6mm}
\end{table}

\begin{table}[t]
\fontsize{10}{12}\selectfont
 \setlength{\tabcolsep}{2.0mm}
  \centering
    \begin{tabular}{l c}
        \toprule 
        {\bf Method}  & {\bf Overall Quality} \\
        \midrule
        E2E  &  24.7\%   \\
        Ours  &  51.7\%  \\
        \midrule
        PlanCoT & 17.8\% \\
        Ours & 70.0\%
       \\
        \bottomrule
    \end{tabular}
    \vspace{2mm}
    \caption{
    Pairwise human evaluations on overall quality.  We report percentage of times the results considered as better.
  }
  \label{tab:human_results_300words}
  \vspace{-6mm}
\end{table}

\begin{figure}[t]
    \centering
    \includegraphics[scale=0.36]{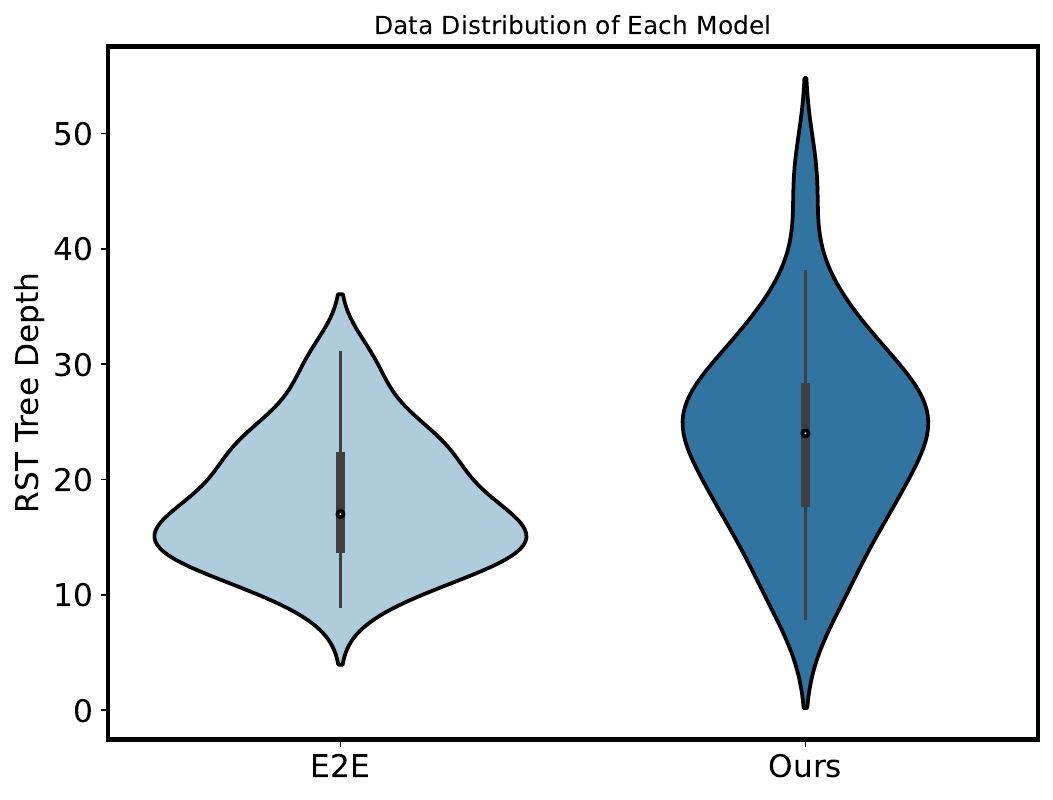}
    \vspace{-3mm}
    \captionof{figure}{ 
    Distribution of RST tree depth of generated arguments under length constraints.
    }
    \vspace{-4mm}
    \label{fig:rst_depth_300words}
\end{figure}


\begin{figure*}[t]
    \centering
    \includegraphics[scale=0.5]{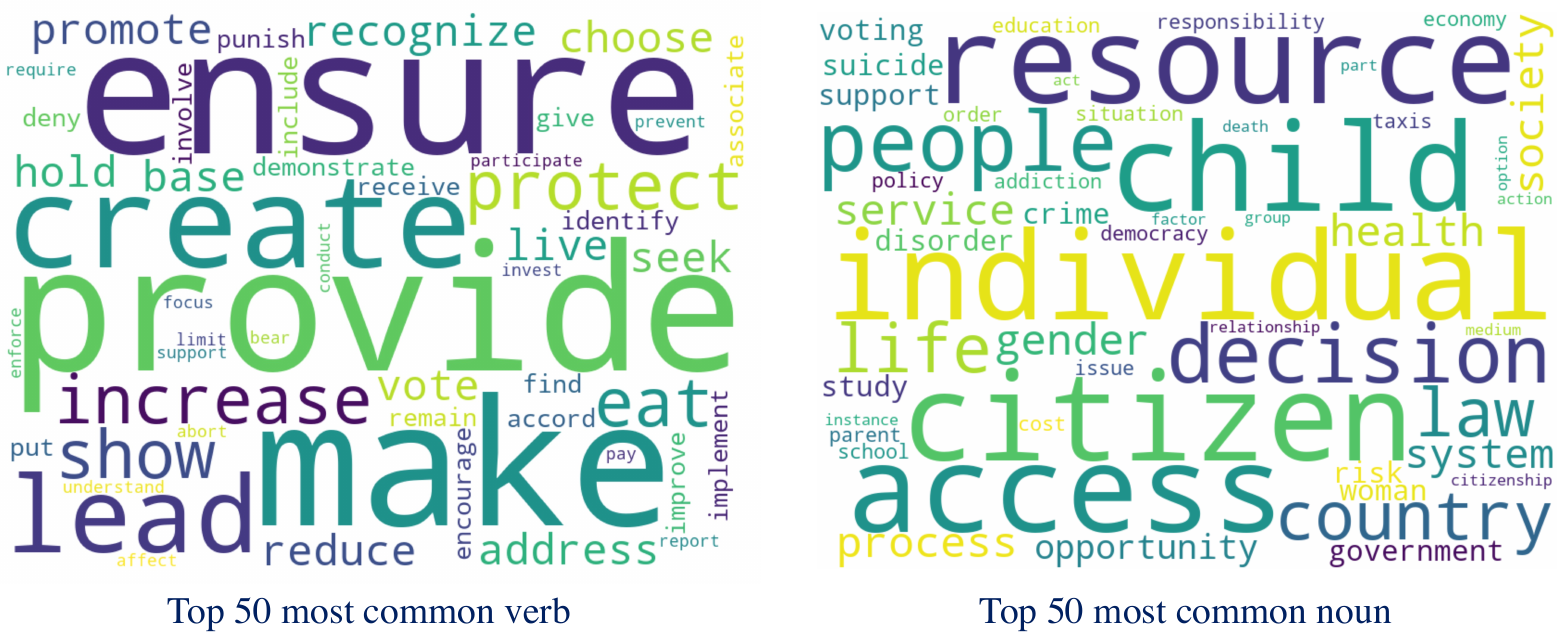}
    \captionof{figure}{ 
    The top 50 most common verbs and nouns in the arguments generated by our method.
    }
    \vspace{-3mm}
    \label{fig:wordcloud}
\end{figure*}

\begin{figure}[t]
    \def\arraystretch{1.5}
	\fontsize{8}{9}\selectfont
     \hspace{-2mm}
	\setlength{\tabcolsep}{0.8mm}
	\centering
	\begin{tabular}{|p{76mm}|}
    \hline

    Background:\\
    Given a statement: [{\_proposition\_}]\\
    We want to generate a counterargument to refute the statement.\\
    -----------------------------------------\\
    Task:\\
    Given a claim and a reasoning of the counterargument, we want to generate a short and brief concession to deal with potential dissenters and predict problems that might weaken the claim and reasoning.\\
    Claim: [{\_claim\_}]\\
    Reasoning: [{\_reason\_}]\\
    Note: the goal of the concession is not to weaken the claim and reasoning, but to strengthen them by demonstrating that you have considered multiple perspectives and can respond to opposing viewpoints effectively. By acknowledging valid points from the opposing side, you build credibility and show that you are open to a fair and balanced discussion. A potential solution might be included in the concession. The concession should be in one sentence.\\
    Concession:

    \\
	
    \hline
	\end{tabular}
    \vspace{-3mm}
\caption{Prompt for concession generation. 
} 
\label{fig:prompt_concession}
\vspace{-2mm}
\end{figure}

\begin{figure}[t]
    \def\arraystretch{1.5}
	\fontsize{8}{9}\selectfont
     \hspace{-2mm}
	\setlength{\tabcolsep}{0.8mm}
	\centering
	\begin{tabular}{|p{76mm}|}
    \hline

    Task:\\
    Given a proposition: {\_proposition\_}\\
    We want to generate a coherent and persuasive counterargument to refute the proposition. You should first generate a brief plan, and then produce the counterargument based on the plan. The output should be in the format of:\\
    Plan:{the generated plan here}\\
    Counterargument: {the generated counterargument here}

    \\
	
    \hline
	\end{tabular}
    \vspace{-3mm}
\caption{Prompt for PlanCoT. 
} 
\label{fig:prompt_cot}
\vspace{-3mm}
\end{figure}

\begin{figure}[t]
    \def\arraystretch{1.5}
	\fontsize{8}{9}\selectfont
     \hspace{-2mm}
	\setlength{\tabcolsep}{0.8mm}
	\centering
	\begin{tabular}{|p{76mm}|}
    \hline

    Proposition: [{\_proposition\_}]\\
    Counterargument: [{\_argument\_}]\\
    Task: Assume you are a professional writer. Given the statement and the counterargument aiming to refute the statement, please evaluate the counterargument based on the following aspects:\\
    * Relevance: The counterargument should directly address the main claim or statement being challenged, rather than introducing tangential or irrelevant points;\\
    * Correct Stance: The counterargument should have a different stance, in order to refute the given statement;\\
    * Logical consistency: The counterargument should be logically consistent and not contain any contradictions or fallacies that weaken its credibility;\\
    * Coherence of structure: The counterargument should have a clear and well-structured progression, with each idea logically flowing from the one before it;\\
    * Persuasiveness: The counterargument should be strong enough to successfully challenge the original statement, which means it should be backed up by solid evidence, clear reasoning, and logical consistency.\\
    Please return a one-paragraph suggestion on how to improve it based on the above criteria.\\Suggestions:

    \\
	
    \hline
	\end{tabular}
    \vspace{-3mm}
\caption{Prompt for evaluation agent in the refinement module. 
} 
\label{fig:prompt_refine_evaluator}
\vspace{-3mm}
\end{figure}

\begin{figure}[t]
    \def\arraystretch{1.5}
	\fontsize{8}{9}\selectfont
     \hspace{-2mm}
	\setlength{\tabcolsep}{0.8mm}
	\centering
	\begin{tabular}{|p{76mm}|}
    \hline

    Proposition: [{\_proposition\_}]\\
    Counterargument: [{\_argument\_}]\\
    Feedback: [{\_feedback\_}]\\
    Task: Assume you are a professional writer. Given the statement, a counterargument on the opposing stance to refute the statement, and a feedback on how to improve the counterargument. Please revise the counterargument based on the feedback.\\
    Revised counterargument:

    \\
	
    \hline
	\end{tabular}
    \vspace{-3mm}
\caption{Prompt for refinement agent in the refinement module. 
} 
\label{fig:prompt_refine_revisor}
\vspace{-3mm}
\end{figure}

\begin{figure*}[t]
    \def\arraystretch{1.5}
	\fontsize{8}{9}\selectfont
     \hspace{-2mm}
	\centering
	\begin{tabular}{|p{150mm}|}
    \hline
    \textbf{Coherence:}\\
    \hdashline
    * Clarity: The counterargument should be expressed clearly, with a well-defined structure that is easy to follow. Ambiguity or vagueness can detract from the argument's coherence;\\
    * Relevance: The counterargument should directly address the proposition and stay focused on the topic. Irrelevant points or anecdotal evidence can detract from its coherence;\\
    * Validity of reasoning: Evaluate the clarity and coherence of the counterargument's reasoning. Is the line of reasoning easy to follow? Does it present a clear cause-and-effect relationship or logical progression? A well-structured and coherent counterargument should present a logical flow of ideas;\\
    * Logical consistency: Assess the counterargument for internal consistency. It should not contain any contradictory statements or logical fallacies that undermine its coherence. Look for logical connections and coherence between the counterargument's claims, evidence, and reasoning
    \\
    \hline\hline
    \textbf{Persuasion:}\\
    \hdashline
    * Language and rhetoric: The counterargument should effectively use persuasive language and rhetoric techniques, such as appeals to logic, emotions, or ethics, to sway the reader's opinion;\\
    * Addressing opposing viewpoints: A persuasive counterargument should acknowledge and address the main points of the given proposition, demonstrating an understanding of the opposing view and refuting it effectively;\\
    * Credibility of evidence: The counterargument should be supported by credible evidence or sources. Unsupported claims or anecdotal evidence will not be as persuasive as a counterargument based on solid evidence;\\
    * Overall effectiveness: The counterargument should effectively challenge the initial proposition and provide a convincing alternative viewpoint, and is likely to persuade the reader to reconsider their initial position.

    \\
	
    \hline
	\end{tabular}
    \vspace{-3mm}
\caption{Specific criteria for GPT4-based automatic evaluation on coherence and persuasion.} 
\label{fig:prompt_auto_eval}
\vspace{-3mm}
\end{figure*}

\section{Automatic Evaluation with GPT4}
\label{sec:gpt4_eval}
For GPT4-based automatic evaluation as described in Section~\ref{sec:gpt4_auto_eval}, the prompts used for evaluation are designed 
with specific task instructions and a comprehensive 
list of detailed criteria, as in Figure~\ref{fig:gpt4_eval}.
We present the detailed criteria for coherence and persuasion in Figure~\ref{fig:prompt_auto_eval}. The description of the criteria is concatenated with the task instruction as the final prompt.

In our initial experiments, we find that the GPT-4 predictions are not very stable. This observation is consistent with prior work~\cite{shen2023large,wang2023large}. To mitigate this problem, instead of directly prompting GPT-4 to predict a score, we first ask it to provide a detailed rationale on evaluation, and then predict the score, which is similar to chain-of-thought prompting~\cite{wei2022chain}. By this strategy, we find the stability of predictions improves by a large margin.  

Another observation is that when evaluating coherence, GPT-4 evaluator tends to prefer shorter results or longer outputs with multiple paragraphs. 
This is a possible reason that in Figure~\ref{fig:gpt4_pred_res}, our decomposed generation receives a lower score on coherence compared with E2E. However, our model with refinement achieves a higher coherence score, as during the refinement process, the refinement agent tends to produce longer outputs with more paragraphs (e.g., average 2.44 paragraphs for ours v.s. 2.12 paragraphs for ours w/o refine). We leave further analysis to future work.

\section{Additional Experiments with Supervised Models}
Previous work on argument generation mainly utilizes smaller models with supervised method~\cite{hua-etal-2019-argument-generation,hua-wang-2018-neural}. In this work, we do not include methods before GPT due to two reasons: (1) We focus on zero-shot argument generation, while the previous method (e.g., BART, T5) requires supervised training; (2) The significant difference in model scales compared with previous methods would lead to an unfair comparison. For the reference, we include two strong (supervised) planning-based methods on long-form text generation: BowPlan~\cite{kang-hovy-2020-plan} and ContentPlan~\cite{hua-wang-2020-pair}. Specifically, BowPlan is a Seq2seq model that predicts keywords as the global plan to guide the surface generation. ContentPlan is a two-step generation method where a planner first produces high-level plans, and then a generator consumes the plans and generates final outputs. We adopt the CMV dataset from ~\citet{hua-wang-2020-pair} for model training, and ensure there is no overlap between the training set and test set used in our paper. We use Bart-large as the base model. The automatic results evaluated by GPT4 are shown in Table~\ref{tab:supervised_baseline}. Our model significantly outperforms both baselines and generates more persuasive and coherent outputs, demonstrating our model effectiveness for argument generation.

\begin{table}[t]
\fontsize{10}{12}\selectfont
 \setlength{\tabcolsep}{2.0mm}
  \centering
    \begin{tabular}{l c c}
        \toprule 
        {\bf Method}  & {\bf Persuasion} & {\bf Coherence} \\
        \midrule
        Ours  & 3.72 & 4.27   \\
        BowPlan  &  1.63 & 1.79 \\
        ContentPlan & 1.23&1.45
        \\
        \bottomrule
    \end{tabular}
    \vspace{2mm}
    \caption{
    Model results compared with supervised baselines.
  }
  \label{tab:supervised_baseline}
  \vspace{-6mm}
\end{table}

\section{Analysis on Model Performance Under Length Constraints}
\label{sec:length_analysis}
In our experiments, we do not impose specific length constraints on the generated outputs due to the open-ended nature of the argument generation task. Our model variants with decomposed generation tend to produce longer outputs than baseline methods (i.e., on average 310 words for our model v.s. 120 words for E2E). In this section, we further analyze the influence of introducing length constraints by specifying the desired output length. In particular, we explicitly include ``\textit{counterargument in around 300 words}'' in the prompts for all methods and further analyze the results. By doing so, the average output lengths of our model, E2E, and PlanCoT change to around 378, 303, and 240 words, respectively.

The GPT4-based automatic scores are shown in Table~\ref{tab:automatic_results_300words}. As can be seen, our model outperforms both E2E and PlanCoT in terms of coherence and persuasion scores. The results are consistent  with the observations where no length constraints are imposed. These findings confirm that our approach, with decomposed generation and subsequent refinement, is highly effective in producing high-quality outputs.

We then conduct human evaluations of the model outputs using pairwise comparison. Specifically, we ask three human annotators to rate the overall quality of the outputs. Given an input, they are shown two outputs, with one generated by our model and one generated by a baseline method, presented in random order. They are asked to select which one is better, and ties are allowed if the outputs are not distinguishable. The results are summarized in Table~\ref{tab:human_results_300words}. Our model results are considered as better with more times than both baselines, underscoring our model effectiveness even when operating under length constraints.

\smallskip
\noindent\textbf{Discourse Diversity.}
We also analyze the output diversity by visualizing the RST trees of outputs. As shown in Figure~\ref{fig:rst_depth_300words}, although E2E generates significantly longer outputs under length constraints, the distribution of RST tree depth is still less diverse compared to our model. This further demonstrate our model ability to produce outputs with greater diversity in rhetorical structure.

\section{Visualization of Common Words}
\label{sec:word_cloud}
We present the visualization top 50 most common verbs and nouns of our model-generated results with word cloud, as displayed in Figure~\ref{fig:wordcloud}. The larger word means the higher frequency. Overall, we can see that our model can generate quite diverse surface formats in the results. Notably, most nouns are policy-relevant, and this is because our dataset is in the politics and policy domains.

\section{Details for Human Evaluation}
\label{sec:human_eval_details}
Three human judges were hired to conduct the evaluation, all of whom are proficient English speakers with at least a Bachelor's degree. We presented 30 random samples for human evaluation. To minimize bias, we anonymized the model names and presented the outputs in a random order. The annotation process spanned two days, allowing all participants enough time to complete their evaluations. 
We evaluate model outputs on the following aspects:

\begin{itemize}[noitemsep,nolistsep,wide]
    \item {\textbf{Appropriateness}}: 
    whether the content is expressed clearly, without ambiguity, vagueness, or grammatical errors; whether it has a good overall structure and strong readability, and the overall logic is smooth, consistent, and complete, with no internal contradictions or incoherence, and the main conclusion can be strongly supported by sub-arguments;
    \item {\bf Content Richness:} whether the output is abundant, with sufficient points and evidence to effectively understand and refute the original input perspective; whether the expression is diverse, with varied diction and different forms of argumentation;
    \item {\bf Overall Quality:} this is a general assessment that whether you think the output ranks higher than all other candidates.

Beside the above ranking-based aspects, human annotators are asked to evaluate the \textbf{Validity} of each output independently, determining whether it constitutes a high-quality and valid argument that effectively refutes the input proposition.

\end{itemize}

\section{Additional Sample Outputs}
\label{sec:addition_samples}
We present additional examples with different model outputs from Figure~\ref{fig:additonal_samples1} to Figure~\ref{fig:additonal_samples4}.

\input{appendix_samples}

%% file: appendix_samples.tex
\begin{figure*}[t]
    \bgroup
    \def\arraystretch{1.8}
	\fontsize{8}{9}\selectfont
     \hspace{-1mm}
	\setlength{\tabcolsep}{0.8mm}
	\centering
	\begin{tabular}{p{155mm}}
	\toprule
	\textbf{Proposition}: The fact that voting isn't mandatory is a good thing. \\
	\midrule	
	\textsc{\textbf{E2E}}: Mandatory voting could help to increase voter turnout, which is important for achieving a fair and representative democracy. Low voter turnout can lead to a government that does not accurately reflect the interests of the people, and can result in policies that are not beneficial for all citizens. Mandatory voting could also help to ensure that all voices are heard and that all perspectives are taken into consideration.  \\
	\hdashline
        \textbf{PlanCoT}: The fact that voting isn't mandatory may lead to a decrease in voter turnout, which can have negative effects on our democracy. Mandatory voting can ensure that all citizens take part in the democratic process, and that no one’s opinion is silenced. Furthermore, the government has a responsibility to ensure that all citizens are informed about the importance of voting and the issues at stake, which can encourage more citizens to vote. Ultimately, voting is a crucial part of any democracy, and making it mandatory ensures that everyone has a say in the political process. \\
        \hdashline
        \textbf{$\text{Ours}_\text{w/o Refine}$}: The fact that voting isn't mandatory is not a good thing. Mandatory voting should be implemented because it is a civic duty that ensures that the people's voices are heard and that their opinions are represented in government. It is a fundamental part of democracy and is an essential way to ensure that citizens are engaged in the electoral process, their voices are taken into consideration when decisions are made, and that the government truly reflects the will of the people. Furthermore, mandatory voting can help to reduce the influence of money in politics by ensuring that all citizens are represented and can participate in the democratic process. 
        While it is true that mandatory voting can ensure that the voices of all citizens are heard, it is important to consider how to make the process accessible and fair for those who are unable to vote due to physical or financial limitations. Ultimately, mandatory voting is a way to guarantee that all citizens have an equal say in the political process and that the government is truly representative of the people's wishes.\\
       
        \hdashline
	{\textbf{Ours}}: The fact that voting isn't mandatory is not a good thing. Mandatory voting should be implemented because it is a fundamental part of democracy and is an essential way to ensure that citizens are engaged in the electoral process and their voices are taken into consideration when decisions are made. By requiring citizens to vote, it can help to reduce the influence of money in politics, as more people are participating in the democratic process and their opinions are being represented. Additionally, mandatory voting can help to ensure that the government truly reflects the will of the people, as all citizens have an equal say in the political process. \\
    \vspace{-5mm}
    However, it is important to consider how to make the process accessible and fair for those who are unable to vote due to physical or financial limitations. For example, measures such as providing transportation to and from polling stations, offering absentee ballots, and introducing flexible voting hours can help to make the process easier and more accessible for those who have difficulty voting in person. Additionally, lawmakers should focus on creating laws that promote fairness and accessibility for all citizens, regardless of their socio-economic status. \\
    \vspace{-5mm}
    Overall, mandatory voting is a way to guarantee that all citizens have a voice in the political process and to ensure that the government is truly representative of the people's wishes. By implementing measures that make the voting process easier and more accessible for all citizens, mandatory voting can help ensure that everyone is given an equal chance to participate and have their voices heard.\\
    \bottomrule
	\end{tabular}
    \vspace{-2mm}
\caption{Sample output of different model outputs. 
} 
\label{fig:additonal_samples1}
\egroup
\vspace{-2mm}
\end{figure*}

\begin{figure*}[t]
    \bgroup
    \def\arraystretch{1.8}
	\fontsize{8}{9}\selectfont
     \hspace{-1mm}
	\setlength{\tabcolsep}{0.8mm}
	\centering
	\begin{tabular}{p{155mm}}
	\toprule
	\textbf{Proposition}: The increase in gender and sexuality vocabulary is not needed and ultimately hurts social progress on the topic. \\
	\midrule	
	\textsc{\textbf{E2E}}: The increase in gender and sexuality vocabulary is absolutely necessary and is an important step in helping to advance social progress on the topic. 
    By expanding our language and terminology to accurately reflect the diversity of gender and sexuality, we are creating a more inclusive environment 
    that allows people to identify and express themselves authentically. This in turn can lead to greater acceptance, understanding, and respect of all gender and 
    sexuality identities. Additionally, the use of more specific language can lead to better and more effective conversation about the topic, 
    allowing for deeper understanding and openness to different perspectives.  \\
	\hdashline
        \textbf{PlanCoT}:  
        The increase in gender and sexuality vocabulary is actually essential for social progress on the topic. It allows us to be more precise in our language, 
        which in turn allows us to better understand each other and progress in our discussions. For example, the introduction of terms such as non-binary, gender fluid, 
        and genderqueer has been crucial in allowing us to move away from the binary of male and female, and to recognize and appreciate the spectrum of gender identities and expressions. This has led to more inclusive and understanding conversations, and a greater respect for gender and sexual minorities.\\Furthermore, the increased language can help to empower individuals, as it provides them with the words to express and identify their own gender and sexuality. This can be especially important for those who have felt invisible in the past, and it can create a sense of connection and understanding amongst members of the community.\\Though there may be some drawbacks to increased language, such as confusion or misinterpretation, these drawbacks are far outweighed by the benefits. As language evolves, so do our understandings of gender and sexuality, and this is essential for making progress on the topic.
        
        \\
        \hdashline
        \textbf{$\text{Ours}_\text{w/o Refine}$}:
        
        The given proposition that the increase in gender and sexuality vocabulary is not needed and ultimately hurts social progress on the topic is false. An increase in gender and sexuality vocabulary is crucial to promoting social progress on gender and sexuality topics, as it allows people to more accurately and effectively communicate their ideas and identities. By introducing new terms and expanding existing ones, individuals can more accurately and effectively communicate their thoughts, feelings, and identities. This helps to eliminate misunderstandings and create a more inclusive and open dialogue. For example, by introducing terms such as “non-binary” and “genderqueer”, individuals who don’t identify with traditional gender labels are able to express their gender identity in a more accurate way. This helps to create a more inclusive and accepting environment for people to discuss gender and sexuality issues. Furthermore, the use of new and more precise language allows individuals to better explain and articulate their thoughts and feelings, which can lead to more meaningful conversations and a better understanding of the issues at hand. By increasing gender and sexuality vocabulary, we can create a more open and 
accepting dialogue and move closer to achieving social progress on these issues. \\However, it is also important to recognize that some people may find it difficult to understand and embrace these new terms. 
Thus, it is important to create accessible resources that explain these terms in a non-threatening and easy to understand way. 
This can help to ensure that individuals have the tools they need to understand and accept the new terms, and to participate in meaningful conversations 
about gender and sexuality issues.\\In conclusion, an increase in gender and sexuality vocabulary is essential to fostering social progress on gender and sexuality topics. 
It is a critical tool for allowing individuals to more accurately and effectively communicate their ideas and identities, which can lead to more meaningful 
conversations and a better understanding of the issues at hand. Therefore, it is clear that the increase in gender and sexuality vocabulary is not only needed, 
but also beneficial in promoting social progress on the topic.
        
        \\
       
        \hdashline
	{\textbf{Ours}}:
 The given proposition that the increase in gender and sexuality vocabulary is not needed and ultimately hurts social progress on the topic is false. 
An increase in gender and sexuality vocabulary is essential to promoting social progress on gender and sexuality topics, as it allows people to more accurately 
and effectively communicate their ideas and identities. By introducing new terms and expanding existing ones, individuals can more accurately and effectively 
communicate their thoughts, feelings, and identities, and eliminate misunderstandings. This can lead to more meaningful conversations and a better understanding 
of the issues at hand. For example, by introducing terms such as “non-binary” and “genderqueer”, individuals who don’t identify with traditional gender labels are 
able to express their gender identity in a more accurate way. This helps to create a more inclusive and accepting environment for people to discuss gender and 
sexuality issues. Additionally, studies have found that the more precise language used to discuss gender and sexuality topics is associated with higher levels of 
acceptance of diverse gender identities and sexual orientations. \\Furthermore, the use of new and more precise language allows individuals to better explain and articulate their thoughts and feelings, which can lead to more 
meaningful conversations and a better understanding of the issues at hand. By increasing gender and sexuality vocabulary, we can create a more open 
and accepting dialogue and move closer to achieving social progress on these issues.   However, it is also important to recognize that some people 
may find it difficult to understand and embrace these new terms. Thus, it is important to create accessible resources that explain these terms in a 
non-threatening and easy to understand way, like glossaries or guides. This can help to ensure that individuals have the tools they need to understand 
and accept the new terms, and to participate in meaningful conversations about gender and sexuality issues. \\In conclusion, an increase in gender and sexuality vocabulary is essential to fostering social progress on gender and sexuality topics. 
It is a critical tool for allowing individuals to more accurately and effectively communicate their ideas and identities, 
which can lead to more meaningful conversations and a better understanding of the issues at hand. Furthermore, the evidence shows that the more precise language 
used to discuss gender and sexuality topics is associated with higher levels of acceptance of diverse gender identities and sexual orientations. 
Therefore, it is clear that the increase in gender and sexuality vocabulary is not only needed, but also beneficial in promoting social progress on the topic.
 
 \\
    \bottomrule
	\end{tabular}
    \vspace{-2mm}
\caption{Sample output of different model outputs. 
} 
\label{fig:additonal_samples2}
\egroup
\vspace{-2mm}
\end{figure*}

\begin{figure*}[t]
    \bgroup
    \def\arraystretch{1.8}
	\fontsize{8}{9}\selectfont
     \hspace{-1mm}
	\setlength{\tabcolsep}{0.8mm}
	\centering
	\begin{tabular}{p{155mm}}
	\toprule
	\textbf{Proposition}: I don't think the duty of child raising should belong to the biological parents. \\
	\midrule	
	\textsc{\textbf{E2E}}: 
 While it is true that in some cases, parents may not be the best suited to raise a child, it is important to remember that biological parents are often the most prepared to do so. 
They typically have a strong bond with their child due to their genetic connection and can provide the necessary emotional, physical, and financial support for their child's development and growth. 
Additionally, biological parents often have a better understanding of their child's specific needs and can make decisions that are in the best interest of their child. 
Thus, while it is important to consider alternative options for child-raising, it is generally in the best interest of the child to be raised by their biological parents.
 
 \\
	\hdashline
        \textbf{PlanCoT}: 
        
        Biological parents have a unique connection to their children that cannot be replicated by anyone else. They understand their children better and are better equipped to provide the emotional support 
and guidance needed to raise healthy, well-adjusted children. When the duty of child raising is not handled by the biological parents, there might be a lack of understanding and emotional connection, 
which can lead to the child feeling neglected or not understood. This, in turn, can have a detrimental effect on the child's development and well-being. 
Therefore, it is important that the duty of child raising still belongs to the biological parents, as they are in the best position to provide the love and support needed to raise a healthy and happy child.
        \\
        \hdashline
        \textbf{$\text{Ours}_\text{w/o Refine}$}:
       I strongly disagree with the proposition that the duty of child raising should not belong to the biological parents. The biological parents are uniquely suited to provide the best possible upbringing 
for their children, and as such, it is their responsibility and privilege to fulfill the duty of child raising. \\The biological parents have a unique understanding of their children's physical and mental needs, which enables them to provide the best possible care. For example, 
parents will be able to recognize the signs of illness or distress in their children quicker than anyone else, and be able to provide the right type of care in a timely manner. 
Furthermore, parents are in the best position to provide the emotional and psychological support that children need to grow and develop. Parents will be able to tailor the type 
of guidance and discipline to the individual needs of their children, and provide an environment that will best nurture them and help them to reach their full potential. 
Thus, it is the unique privilege and responsibility of the biological parents to fulfill the duty of child raising.\\That said, I acknowledge that there may be cases in which non-biological parents may provide similar or even better care for a child, 
and in those cases, it may be appropriate to entrust the duty of child raising to them. This does not, however, negate the fact that, in most cases, it is the biological parents 
who are best suited to provide the best possible upbringing for their children, and therefore it is their responsibility and privilege to fulfill the duty of child raising.\\
        \hdashline
	{\textbf{Ours}}:
 I strongly disagree with the proposition that the duty of child raising should not belong to the biological parents. The biological parents are uniquely suited to provide the best possible upbringing 
for their children, and as such, it is their responsibility and privilege to fulfill the duty of child raising.  The biological parents have a unique understanding of their children's physical and mental needs 
that enables them to provide the best possible care. For example, parents are more likely to recognize signs of distress in their children quicker than anyone else due to their familiarity with their children's
 behaviors, and they can provide the right type of care in a timely manner. Furthermore, parents are in the best position to provide the emotional and psychological support that children need to grow and 
develop. As such, parents are able to tailor the type of guidance and discipline to the individual needs of their children, and provide an environment that will best nurture them and help them to 
reach their full potential. \\To further illustrate this point, research has shown that children who grow up with both biological parents experience better physical and mental health outcomes, higher levels of educational achievement,
 and better economic success later in life. This is further proof of the unique role that biological parents have in providing the best possible upbringing for their children.  
Thus, it is the unique privilege and responsibility of the biological parents to fulfill the duty of child raising. That said, I acknowledge that there may be cases in which non-biological parents 
may provide similar or even better care for a child, and in those cases, it may be appropriate to entrust the duty of child raising to them. Nonetheless, in most cases, it is the biological parents 
who are best suited to provide the best possible upbringing for their children, and therefore it is their responsibility and privilege to fulfill the duty of child raising.
 
 \\
    \bottomrule
	\end{tabular}
    \vspace{-2mm}
\caption{Sample output of different model outputs. 
} 
\label{fig:additonal_samples3}
\egroup
\vspace{-2mm}
\end{figure*}

\begin{figure*}[t]
    \bgroup
    \def\arraystretch{1.8}
	\fontsize{8}{9}\selectfont
     \hspace{-1mm}
	\setlength{\tabcolsep}{0.8mm}
	\centering
	\begin{tabular}{p{155mm}}
	\toprule
	\textbf{Proposition}: There shouldn't be a death penalty for any crime but there should be a death option. \\
	\midrule	
	\textsc{\textbf{E2E}}: The death penalty should remain as a punishment option for some of the most serious crimes, as it serves as a deterrent to potential offenders and sends a clear message that certain actions 
will not be tolerated. By introducing the death option, we are sending the wrong message to would-be criminals, and it would encourage more people to commit serious crimes. 
Furthermore, the death option is not an effective way of achieving justice, as it does not allow for the possibility of rehabilitation or redemption for those who have committed a crime.  \\
	\hdashline
        \textbf{PlanCoT}: The death penalty should not be an option for any crime. Firstly, it is uncertain whether it is an effective deterrent for crime. Studies have shown that the death penalty does not 
have a greater effect than other punishments, such as life imprisonment, in reducing crime rates. Secondly, there is a risk of wrongful executions, as mistakes can be made in the judicial process. 
Finally, while it may be argued that for some serious crimes imprisonment is not an appropriate punishment, attention should instead be directed to reforming the legal system to provide alternative forms 
of justice. \\
        \hdashline
        \textbf{$\text{Ours}_\text{w/o Refine}$}: 
        The death penalty is an essential part of any justice system, and should remain as a punishment for serious crimes. There is no place for a 'death option' 
which would undermine the seriousness of serious criminal acts. The death penalty provides a clear and unambiguous deterrent for serious criminal acts, and ensures that those who commit such 
acts are held accountable for their actions. It also serves as a powerful symbol of the justice system’s commitment to protecting society from the most serious offenders. Without the death penalty, 
there is a risk that offenders could escape appropriate punishment for their crimes. The death option proposed in the original statement would undermine the seriousness of serious criminal acts 
and create a loophole for those who commit them to escape punishment. It would also reduce the deterrent effect of the death penalty, as offenders would know that they could opt for the death 
option and avoid the full consequences of their actions. This could lead to more crime and a less safe society.\\While we should consider alternative sentencing options that provide a fair punishment while also protecting society from further harm, the death penalty is an important part of any justice system. 
It is essential that those who commit serious crimes are held accountable for their actions, and the death penalty serves as a powerful signal of the justice system’s commitment to upholding the law. 
There is no place for a ‘death option’ which would undermine the seriousness of serious criminal acts and create a loophole for those who commit them to escape punishment.
 Therefore, the death penalty should remain as a punishment for serious crimes, and any attempt to introduce a ‘death option’ should be strongly opposed.\\
       
        \hdashline
	{\textbf{Ours}}:The death penalty is an essential part of any justice system, and should remain as a punishment for serious crimes. It serves as a powerful symbol of the justice system's commitment 
to protecting society from the most serious offenders, and provides a clear and unambiguous deterrent for serious criminal acts. Furthermore, the death penalty ensures that those who commit 
such acts are held accountable for their actions. Introducing a 'death option' would undermine the seriousness of serious criminal acts and create a loophole for those who commit them to escape punishment. 
It would also reduce the deterrent effect of the death penalty, as offenders would know that they could opt for the death option and avoid the full consequences of their actions. 
This could lead to more crime and a less safe society. \\Moreover, the death option proposed could also have a negative impact on society by providing less of an incentive for rehabilitation. Offenders may become less likely to seek 
rehabilitation if they know that they can opt for the death option and avoid the full consequences of their actions. This could lead to further criminal activity and a less safe society.\\Therefore, while we should consider alternative sentencing options that provide a fair punishment while also protecting society from further harm, the death penalty 
is an important part of any justice system. It is essential that those who commit serious crimes are held accountable for their actions, and the death penalty serves as a 
powerful signal of the justice system’s commitment to upholding the law. There is no place for a ‘death option’ which would undermine the seriousness of serious criminal acts 
and create a loophole for those who commit them to escape punishment. Therefore, the death penalty should remain as a punishment for serious crimes, and any attempt to 
introduce a ‘death option’ should be strongly opposed. \\
    \bottomrule
	\end{tabular}
    \vspace{-2mm}
\caption{Sample output of different model outputs. 
} 
\label{fig:additonal_samples4}
\egroup
\vspace{-2mm}
\end{figure*}

\input{inputs}

%% file: inputs.tex
\begin{figure*}[t]
    \bgroup
    \def\arraystretch{1.8}
	\fontsize{8}{9}\selectfont
     \hspace{-1mm}
	\setlength{\tabcolsep}{0.8mm}
	\centering
	\begin{tabular}{p{155mm}}
	\toprule
- I think suicide should be a human right

- The US should strictly enforce border security to prevent illegal entry

- Starting a cult should not be protected as free speech

- The majority of the public are too apathetic / uneducated to vote rationally. Democracy is no longer the solution to effective governance.

- Both conservative and liberal economies can work very well, and the devil is in the details.

- The minimum wage should be directly attached to housing costs with low consideration of other factors.

- There is no defensible reason to prefer children of your own genetic material to adopting them.

- Bartenders should be able to refuse liqour service to pregnant women.

- Democracy, as it stands today, is an insufficient form of government and we need to find a replacement

- The U. S. should establish a system whereby other countries can be admitted to the union.

- Employees should not always be blamed for ignoring / inaction on a case of sexual harassment within their company / institution

- The American education would benefit from abolishing public schools and moving to a privatized system, with the government helping those who cannot afford the private schools.

- It is the moral responsibility of a free nation to annihilate those that perpetrate human rights abuses

- Drunk driving should not be a crime itself.

- The increase in gender and sexuality vocabulary is not needed and ultimately hurts social progress on the topic

- Some type of basic understanding exam should be required for anyone who wants to vote.

- I don't think the duty of child raising should belong to the biological parents.

- The whole debate of whether addiction is a choice or disease is pointless and should simply be labeled as bad.

- Poor people must have the choice to be poor, otherwise they are inherently inferior

- CMV :'undocumented immigrant'is a nonsense term from the left and anyone entering the country illegally ( without granted asylum ) should be deported

- Having children is unethical

- There shouldn't be a death penalty for any crime but there should be a death option.

- I Think Groups That Exclude Based on Skin Color or Gender are Supremacy Groups

- People who falsely accuse of rape should get equal prison time as rapists do.

- The fact that voting isn't mandatory is a good thing.

- We should not have laws that govern our own safety

- All bigotry is wrong and immoral, no matter the perpetrator.

- We can get Offended by Media or Ideas ALL we want, but we should NEVER Advocate Suppression of those Ideas or Deletion of that Media

- Within the window that women have to biologically abort, men should be able to financially abort from their paternal responsibilities.

- Having sex with people who are emotionally unavailable due to their commitment to a relationship, knowingly that they are, shouldn't be considered a morally corrupt act.

- basic universal income is useless, due to supply and demand and inflation

- Legal history and politics aside, where you are born has no relevance to citizenship

- Voting Rights Should be Accorded by Residency not Nationality

- There should be 3 and only 3 gendered pronouns.

- Countries should not support eating disorder legislation.

- Selectively breeding animals with genetic defects should be illegal

- The worse the current migrant situation gets, the better the long - term prospects for our immigration system.

- Voting data that segments the voters by gender / race should not be made public.

- It is usually better for governments to offer tax holidays to attract business than to not attract the busines

- Private hospitals should be outlawed.

- Suicide should be legal

- Corporations are inherently evil and society would be better without them.

- Paying taxes cannot be considered virtuous because it is compulsory.

- Women who've been sexually assaulted should take justice into thier own hands.

- Carrying a gun for self - defense as opposed to pepper spray is unnecessary and possibly less safe / effective

- All labels to identify activists or certain groups of people in general ( ex. Feminist, ANTIFA, Alt - Right, Liberal ) are hurting society more than they are helping.

- Torture is sometimes acceptable

- Victimless Crimes Shouldn't Be Illegal

- Monogamy is not the most realistic outcome in many long - term relationships

- Social media sites policing discussions is a mistake \\
    \bottomrule
	\end{tabular}
    \vspace{-2mm}
\caption{List of input propositions sampled from Reddit/CMV dataset~\cite{hua-etal-2021-dyploc,hu-etal-2022-planet}. 
} 
\label{fig:input_query}
\egroup
\vspace{-2mm}
\end{figure*}